\documentclass[onecolumn]{cohere}

\usepackage{lipsum}

\usepackage{amsmath,amsfonts,bm}

\def\eqref#1{equation~\ref{#1}}
\def\1{\bm{1}}

\DeclareMathAlphabet{\mathsfit}{\encodingdefault}{\sfdefault}{m}{sl}
\SetMathAlphabet{\mathsfit}{bold}{\encodingdefault}{\sfdefault}{bx}{n}

\definecolor{orange}{rgb}{1.0, 0.6, 0.1}

\usepackage{stfloats} 
\usepackage{hyperref}
\usepackage{url}
\usepackage{booktabs}
\usepackage{graphicx}
\usepackage{multirow}
\usepackage[export]{adjustbox}
\usepackage{float}
\usepackage{caption}
\usepackage{subcaption}
\usepackage[colorinlistoftodos]{todonotes}
\usepackage{lscape}
\usepackage{rotating}
\usepackage{enumitem}
\usepackage{wrapfig}
\usepackage{tabularx}

\title{When Less is More: \\Investigating Data Pruning for Pretraining LLMs at Scale}

\author{
    name={Max Marion},
    affiliation={Cohere for AI},
    email={maxwell@cohere.com}
}
\author{
    name={Ahmet Üstün},
    affiliation={Cohere for AI},
    email={ahmet@cohere.com}
}
\author{
    name={Luiza Pozzobon},
    affiliation={Cohere for AI},
    email={luiza@cohere.com}
}
\author{
    name={Alex Wang},
    affiliation={Cohere},
    email={alexwang@cohere.com}
}
\author{
    name={Marzieh Fadaee},
    affiliation={Cohere for AI},
    email={marzieh@cohere.com}
}
\author{
    name={Sara Hooker},
    affiliation={Cohere for AI},
    email={sarahooker@cohere.com}
}

\date{\today}
\abstract{Large volumes of text data have contributed significantly to the development of large language models (LLMs) in recent years. This data is typically acquired by scraping the internet, leading to pretraining datasets comprised of noisy web text. To date, efforts to prune these datasets down to a higher quality subset have relied on hand-crafted heuristics encoded as rule-based filters. 
In this work, we take a wider view and explore scalable estimates of data quality that can be used to systematically measure the quality of pretraining data. We perform a rigorous comparison at scale of the simple data quality estimator of perplexity, as well as more sophisticated and computationally intensive estimates of the Error L2-Norm and memorization. These metrics are used to rank and prune pretraining corpora, and we subsequently compare LLMs trained on these pruned datasets.
Surprisingly, we find that the simple technique of perplexity outperforms our more computationally expensive scoring methods. We improve over our no-pruning baseline while training on as little as 30\% of the original training dataset. Our work sets the foundation for unexplored strategies in automatically curating high quality corpora and suggests the majority of pretraining data can be removed while retaining performance. 
}

\begin{document}

\section{Introduction}\label{sec:intro}

A reigning belief in machine learning is that more data leads to better performance. Recent years of progress in scaling large language models (LLMs) have shown strong evidence to support this with remarkable gains in language understanding and generation
capabilities \citep{brown2020language, touvron2023llama, kaplan2020scaling, anil2023palm}.
When training language models, common practice is to use massive datasets such as C4~\citep{raffel2020exploring}, RefinedWeb \citep{penedo2023refinedweb}, and The Pile~\citep{DBLP:journals/corr/abs-2101-00027}. 
These datasets are typically compiled by scraping raw web pages from the internet, leading to a substantial portion of the text being noisy and of low quality \citep{dodge2021documenting, 10.1162/tacl_a_00447, luccioni-viviano-2021-whats}.

Practitioners have established a number of standard filtering techniques to remove low-quality examples from these datasets.
These techniques are predominantly rule-based heuristics: removing documents containing repetitive text \citep{zhang2022opt, raffel2020exploring, rae2022scaling, hernandez2022scaling, penedo2023refinedweb}, special characters, or non-English text \citep{wenzek-etal-2020-ccnet}; ignoring data from a manually curated list of ``blocklist'' websites \citep{dodge2021documenting, rae2022scaling}; or eliminating documents based on certain length thresholds. 
While these hand-curated filters can eliminate certain noisy examples, they are not a substitute for a measure of ``quality'' for individual training examples, for which there are currently no established best practices \citep{mitchell2023measuring}.

\begin{figure}[t]
 \centering 
 \includegraphics[width=0.8\textwidth]{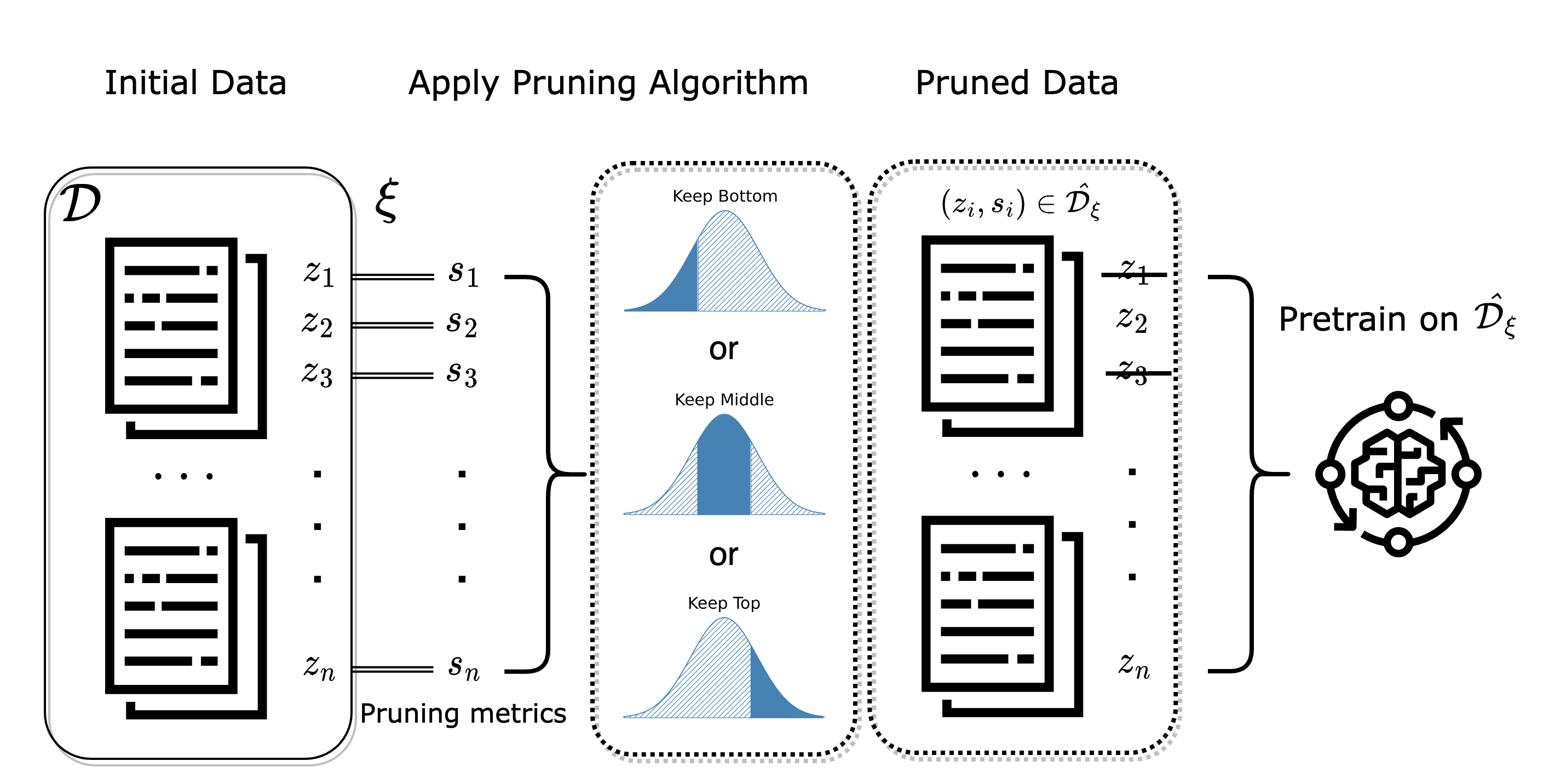}
    \caption{Demonstration of our pruning methodology.For each sequence $z_i$, sized equally as the model's context length, a pruning algorithm $\xi$ generates score $s_i$. We then choose which subset of the distribution of scores to keep: bottom, middle, or top. Finally, a new model is pretrained with the pruned data $\mathcal{\hat{D}}_{\xi}$.}
    \label{fig:distributions}
\end{figure}

In this work, we take a wider view and ask if we can arrive at a rigorous estimator of data quality through \textit{data pruning}. 

Data pruning attempts to isolate a subset of a larger training dataset such that a model trained on said subset preserves or improves performance over a model trained on the full dataset. To date, the majority of work on data pruning has centered on supervised computer vision settings \citep{qin2023infobatch, sorscher2023neural, raju2021accelerating, paul2023deep, he2023largescale}, with far fewer works focusing on language. Those that have either studied the fine-tuning setting, which typically has an order of magnitude less data and thus tolerates more computational complexity \citep{fayyaz2022bert, attendu2023nlu, cao2023instruction} or based their method on hand picking high-quality corpora \citep{gao2021empirical, wenzek-etal-2020-ccnet, brown2020language}.
Specifically, we try to answer the following: \textit{Can we remove the least impactful examples from a pretraining dataset and achieve similar or better performance?} \textit{Do simpler techniques for estimating data quality outperform more sophisticated and computationally expensive methods?} 
\textit{What aspects of training dynamics signal data quality the best?}

We answer these questions by rigorously evaluating three automatic pruning metrics. One simple estimator of quality, perplexity, and two more complex, and EL2N \citep{paul2023deep} memorization factor. 
These methods all rely solely on model outputs and do not require a preselected high-quality dataset. This lack of dependence on human judgments of data quality make them a promising direction for automatic selection of high quality corpora.
We perform extensive experiments evaluating models ranging from 124M to 1.5B parameters across different pretrained corpora. Our contributions are the following:

\begin{enumerate}[topsep=0pt,itemsep=-1ex]
\itemsep0pt
  \item We extensively benchmark data pruning based on perplexity, EL2N, and memorization in the LLM pretraining setting.
  \textbf{Surprisingly, we find the simple technique of ranking examples based on their perplexity outperforms far more complex techniques such as memorization.} A model trained on 50\% of the dataset pruned based on perplexity achieves 1.33\% and 1.77\% improvement over the most performant models pruned to 50\% of the dataset with EL2N and memorization factor respectively. A model trained on 30\% of the dataset pruned with perplexity achieves a 2.1\% and 1.6\% improvement over the most performant models pruned to 30\% of the dataset with EL2N and memorization factor.
  \item To comprehensively cover multiple facets of data pruning, we provide a unified and general framework to identify and treat different data subsets present in a dataset. 
  We compare models trained on datasets pruned to 10, 30, 50, and 70\% of the training set while retaining either the \texttt{bottom}, \texttt{middle}, or \texttt{top} of the pruning scores' distributions. We test seven different reference models across pruning variations, investigating the impact of parameter count, training dataset, and total training steps on the reference models' pruning capabilities.
  Finally, we finetune a selection of our models on six tasks from the GLUE benchmark \citep{wang2019glue} to evaluate the effect of pruning on downstream generalization.
  \item We test our pruning methods at scale, achieving a 1\% improvement in test set perplexity using half of the dataset over a baseline model trained on the entire dataset. We show this scales to 1.5B parameter models, achieving 1.5\% improvement in test set perplexity over a no-pruning baseline of the same size.
\end{enumerate}

\section{Methodology}\label{sec:methodology}

Given a large-scale dataset $\mathcal{D}$, we tokenize all documents and append a special \texttt{\textless eod\textgreater} token to their end. We then concatenate and split them into $n$ sequences $z_i$ of fixed length $t$ equal to the model's context length: $\mathcal{D}=\left\{z_1, \ldots, z_n\right\}$.
Consider the subset of training instances $\mathcal{P}_\xi$ where $\xi$ refers to the algorithm used to select the subset.
We build this subset by computing the pruning score $Score_\xi(z_i)$ for each data point $z_i$.
We then populate $\mathcal{P}_\xi$ with instances that fit our selection criteria:
\begin{equation}
    \label{eq:subset_selction}
    \mathcal{P}_\xi = \{z_i\in\mathcal{D}\ |\ Criteria(Score_\xi(z_i)) \}
\end{equation}
By removing $\mathcal{P}_\xi$ from $\mathcal{D}$, the remaining instances are described as:
\begin{equation}
    \hat{\mathcal{D}}_\xi = \mathcal{D}\setminus \mathcal{P}_\xi
\end{equation}
Our goal is to choose the pruning algorithm $\xi$ such that when training a language model on the remaining subset of training instances, $\hat{\mathcal{D}}_\xi$, the model's performance is not diminished:
\begin{equation}
\mathbb{P}_{\tau}(\mathcal{M}_{\hat{\mathcal{D}}_\xi}) \geq \mathbb{P}_{\tau}(\mathcal{M}_\mathcal{D})
\end{equation}
where $\mathcal{M}_{\hat{\mathcal{D}}_\xi}$
is the model trained on $\hat{\mathcal{D}}_\xi$ 
and $\mathbb{P}_\tau$ is the performance on task $\tau$. 
We explore three metrics, perplexity, Error L2-Norm (EL2N), and memorization which we detail below in Section \ref{sec:pruning_methods}, and evaluate the different ways in which the metric can be employed to determine $\mathcal{P}_\xi$.

In particular, we evaluate different reference models $\Tilde{\mathcal{M}}$ that are used to calculate pruning scores.
Both reference models $\Tilde{\mathcal{M}}$ and trained models $\mathcal{M}$ share the same context length to ensure consistency between the contexts for which pruning metrics are calculated and trained models are trained.

For each metric, we consider three different selection criteria to determine $\mathcal{P}_\xi$ as seen in Equation \ref{eq:subset_selction}: isolating the \texttt{top}, \texttt{middle}, or \texttt{bottom} percentiles of $\mathcal{D}$ as the data to be kept.
We pretrain separate models using these criteria with different percentages of the dataset to understand the dynamics and impact of each pruning metric.
Since the effectiveness of these metrics in this specific context remains uncertain, we opt for these contrasting subsets to clarify the relationship between each metric and the overall model performance.
Figure~\ref{fig:distributions} demonstrates our experimental setup. 
We focus on static pruning, in which data is pruned once before training. This is in contrast to adaptive pruning, in which data is pruned as training is happening, such as in \citep{fayyaz2022bert,park2022active}. 

\subsection{Pruning Methods}\label{sec:pruning_methods}
Here, we briefly describe data pruning algorithms that we benchmark in this work. Our goal is to rigorously compare simple and computationally inexpensive ranking approaches such as \textsc{perplexity} and \textsc{random ranking} 
against more sophisticated and computationally expensive techniques such as \textsc{memorization} scores and \textsc{EL2N}.

\subsubsection{Selection via Perplexity}
\textsc{Perplexity} measures how probable a given piece of text is based on a particular language model.
For each instance $z_i$ in $\mathcal{D}$,
 we compute the perplexity metric as:
\begin{equation}
PPL(z_i) = \exp\big(\frac{1}{|z_i|} \sum_{t_j \in z_i} NLL(t_j)\big)
\end{equation}
where $NLL(t_j)$ is the negative log likelihood of token $t_j$ in sequence $z_i$:
\begin{gather}
\label{eq:nll}
    NLL(t_j)=-\log P(t_j|t_{<j}; \theta )
\end{gather}
A lower perplexity score indicates that the model assigns a high probability to the text.

\subsubsection{Selection via EL2N}
The Error L2-Norm (\textsc{EL2N}) score was originally proposed in a computer vision setting to identify which samples are important for learning \citep{paul2023deep}. 
It measures each sample's importance using the model's early learning signals.
We define the \textsc{EL2N} score on text sequences as the average $L_2$ norm of the error vector, where $\hat{y}_i$ is the reference model's predicted probability distribution over the vocabulary and $y_t$ is the one-hot encoded representation of the ground truth:
\begin{equation}
    \label{eq:el2n}
    \text{EL2N}(z_i) = \frac{1}{t} \sum_{i}^t\| \hat{y}_t - y_t \|_2
\end{equation}
We first evaluate the pruning efficacy of \textsc{EL2N} scores obtained from a single reference model at two different checkpoints, trained on 14\% and 55\% of the training dataset $\mathcal{D}$ corresponding to 250 and 1000 steps respectively, to determine the required number of steps needed before a usable pruning signal emerges.
We then train ten different reference models with different random initializations and average the EL2N score from all ten models to obtain our final EL2N score.
The authors suggest that exhibiting a low \textsc{EL2N} score are typically those the model learns in its early stages of training, likely because they are relatively easier. Inversely, examples with higher \textsc{EL2N} scores are hypothesized to indicate that the model continues to incur a significant loss for them and may require additional iterations to learn.

\subsubsection{Memorization Ranking}
Memorization in language models is a well-studied phenomenon \citep{carlini2023quantifying, carlini2021extracting, biderman2023emergent}. In this work we explore memorization scores applied as a data pruning ranking. 
We use the memorization score as defined by \citet{biderman2023emergent}:
\begin{equation}
score(M,N) = \frac{1}{N} \sum_{i}^{N}1(z_{M+i} = \hat{z}_{M+i})
\end{equation}
where $z$ is a data point, $\hat{z}$ is a sequence of tokens predicted by the reference model, and $1(\cdot)$ is an indicator function.
A reference model is prompted with the first $M$ tokens of a data point $z$ to calculate the memorization score. We then greedily generate $N$ additional tokens, $\hat{z}$. 
The memorization score is the fraction of the $N$ greedily generated tokens ($\hat{z}_{M:M+N}$) that match exactly with the original data point ($z_{M:M+N}$).
For our experiments, $M = N = 32$. We note that the authors did not originally propose this as data pruning metric, but we hypothesize that it can be a valuable ranking to identity examples which require additional learning. We use reference models guaranteed to have seen the full training set to ensure the applicability of memorization scores. 
A high memorization score indicates the model reproduces more of the text verbatim.  

\subsubsection{Random Pruning} We also evaluate a lower bound of expected performance: pruning a random selection of samples. This allows us to ask the question ``are proposed pruning methods any better than a \emph{random guess}?''

\section{Experiments}

\subsection{Model} We train autoregressive decoder-only Transformer models \citep{vaswani2023attention} with a standard language modeling objective. 
Given an input sequence of $z_i=$ $\left[r_1, \cdots, r_t\right]$ 
from training data $\mathcal{D}$, a language model with parameters $\theta$ is trained to minimize the negative log-likelihood loss as defined in Equation \ref{eq:nll}. Our language models follow the traditional GPT-style architecture \citep{radford2018improving}.

While training our models, we use AdamW~\citep{loshchilov2018decoupled} with linear cosine scaling and a batch size of 2048. The 124M parameter models are trained for 8000 steps, which amounts to a total of 33B tokens with a learning rate that linearly increases from 0 to 1.5e-4 over the course of training. This is approximately 4.4 epochs over the unpruned dataset.
We tokenize the data with Byte Pair Encoding \citep{sennrich2016neural} with a vocabulary of 51200.
Due to the memory and computational costs of training 1.5B parameter models, our experiments at this size are trained with a batch size of 512 for 14568 steps.
As such, the models see only 7.6B tokens, equivalent to a single epoch of our unpruned dataset.
The learning rate for 1.5B parameter models linearly increases from 0 to 1.2e-4 over the course of training.
All models use a context window length of 2048.
\subsection{Data}
We use a random sample of the May 2022 snapshot of CommonCrawl\footnote{https://data.commoncrawl.org/} in our experiments. After downsampling the unpruned dataset has 7.6B tokens, about 20\% of the full snapshot. This downsampling is required due to the computational cost of our various ablation experiments, which each require pretraining a new model from random initialization.  This dataset is \textit{prefiltered} using a combination of automatic and hand-crafted filters, as we aim to further improve data quality beyond common rule-based filters. The filters exclude repetitive documents, documents with percentages of special characters, and documents that contain explicit words and toxic text, similar to deduplication steps seen in \cite{taylor2022galactica, kocetkov2022stack}. Our Wikipedia dataset contains 5.3M tokens and only includes English pages. 

\subsection{Ablations} 

\begin{table}
    \centering
    \begin{tabular}{ll}
        \toprule
        Experimental axes  & Choices \\
        \midrule
        Pruning Metric  & Perplexity, EL2N, Memorization \\
        Pct. Data Remaining & 10, 30, 50, 70 \\
        Pruning Subset & Bottom, Middle, Top \\
        Reference Model Size & 124M, 6B, 13B, 52B \\
        Reference Model Epoch Perc. & 14\%, 55\%, 440\%, Full \\
        Reference Model Tr. Data & CC, Wiki, Web-scale\\
        Trained Model Size & 124M, 1.5B \\
        \bottomrule
    \end{tabular}
    \caption{\label{tab:expsetup} Pruning choices explored in the experiments. Under ``Reference Model Training Steps'', ``Full'' refers to the fully trained Cohere LLMs. Under ``Reference Model Training Data'', ``Web-scale'' refers to the significantly larger training datasets used by the Cohere reference models.}
\end{table}

For all techniques, we compare performance when only 10\%, 30\%, 50\%, and 70\% of all data is preserved. We compare retaining the \texttt{top}, \texttt{middle}, and \texttt{bottom} subsets according to the pruning ranking, e.g., when retaining 30\% of the bottom of the pruning metric's distribution over the training set, we calculate the 30th percentile of the pruning metric's distribution and remove all data points with perplexity above it.
When retaining the \texttt{middle} 30\%, we calculate the 35th and 65th percentile and remove all data points above and below those numbers respectively. 
Each ablation study(pruning method, percent data remaining, section of distribution preserved) \textbf{requires training a new model from random initialization}. We train a minimum of nine models with 124M parameters from scratch for each experimental variant.

Table~\ref{tab:expsetup} summarizes the perplexity pruning variations we explore in this paper. 
For perplexity, we use a separate model to compute perplexity from the model trained on the pruned data.
We call models used to compute the perplexity ranking \textit{reference models} and the models trained on the pruned datasets \textit{pruned models}. We conduct a rigorous evaluation of what impacts the quality of the ranking by varying different factors that affect the perplexity distribution:
\begin{enumerate}
    \item \textbf{Reference Model Size} To explore how reference model size impacts the rating quality, we compare perplexity computations using 6B, 13B, and 52B Cohere models trained on full web-scale datasets.
    \item \textbf{Reference Model Training Data} To isolate the impact of training data, we compute perplexity using 124M parameter reference models trained on either CommonCrawl or Wikipedia.
    \item \textbf{Total Reference Model Training Steps} To isolate the impact of early training signals, we compute perplexity and EL2N using 124M parameter models trained on CommonCrawl data for approximately 14\% and 55\% of total training steps. Reference models trained on CommonCrawl are trained on a non-overlapping subset from the CommonCrawl dataset that is pruned and used to train the student model.
\end{enumerate}

\subsection{Evaluation} 
\label{sec:evaluation}

We report perplexity on a test set from the same CommonCrawl snapshot with identical prefiltering as the training data.
This test set contains 266M tokens, equivalent to about 3.5\% of the training set.

We also finetune a subset of our models on six different classification tasks from GLUE \citep{wang2019glue}.We do not prune the task dataset, as our aim is to analyze the pruning methods' effects on pretraining. 
We compare performance after 8000 steps (approximately 4.4 epochs of the pretraining dataset), chosen to compare performance after models have saturated their capacity by training enough steps to plateau on validation metrics.

\section{Results and Discussion}\label{sec:results}

\begin{figure*}[htb]
    \centering   
    \includegraphics[width=0.9\textwidth]{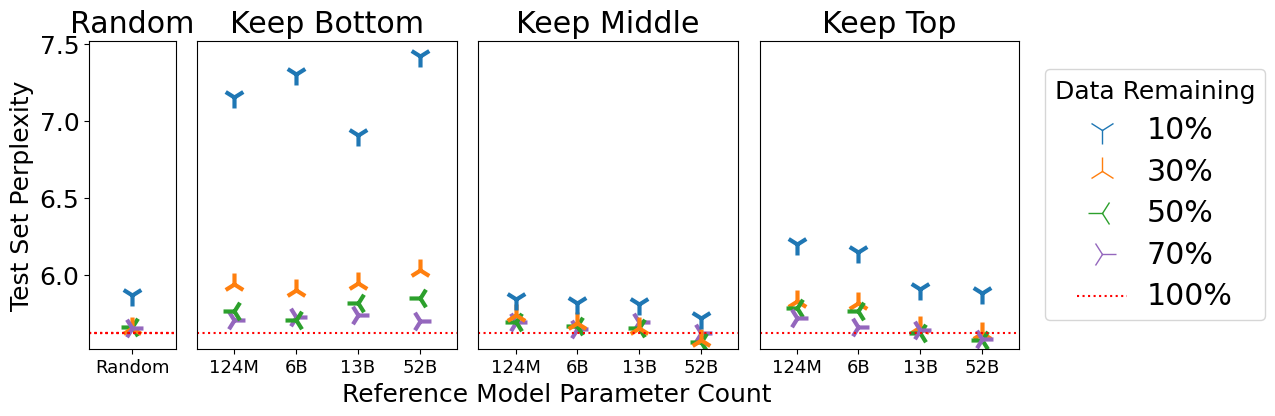}
    \caption{The effect of employing reference models of different sizes on the computation of pruning perplexity scores and its subsequent influence on test set perplexity. The three subset selection approaches for each set of experiments are showcased separately (keeping \texttt{bottom}, \texttt{middle}, or \texttt{top} of the pruning score distribution).}
    \label{fig:impact_of_size}
\end{figure*}

\subsection{Removing Easy Instances Improves Performance}
Though the most competitive variant for each pruning method varies based on the subset of the scoring distribution retained (\texttt{top}, \texttt{middle}, or \texttt{bottom}), we observe a consistent pattern: the highest performant variants are \emph{not} the subsets that correspond to the ``easier'' data.
The interpretation of the term ``easy'' varies according to the measurement employed. 
When employing the \textsc{Perplexity} metric, it refers to the \texttt{bottom} samples with the lowest perplexity.
With the \textsc{EL2N} metric, it also pertains to the \texttt{bottom} samples exhibiting the lowest initial loss. 
In the context of \textsc{memorization}, it relates to the \texttt{top} samples that have been most thoroughly memorized.

Figure \ref{fig:impact_of_size} demonstrates this pattern when using \textsc{Perplexity}. In contrast to the \texttt{middle} or \texttt{top} subsets, the \texttt{bottom} subset has much less variance in results between reference models of varying sizes, indicating the \texttt{bottom} subset may not be suitable for training. 
The \texttt{middle} experiments achieve consistently low test set perplexities for various reference model sizes and pruning ratios. Generally, performance monotonically degrades as the amount of data remaining shrinks - except for the \texttt{middle} subset for the best-performing reference models. In these cases, retaining only 50\% and even 30\% of the dataset outperforms retaining 70\% of the dataset.

\begin{figure}[htb]
\centering
\begin{subfigure}{.5\textwidth}
  \centering
  \includegraphics[width=.7\linewidth]{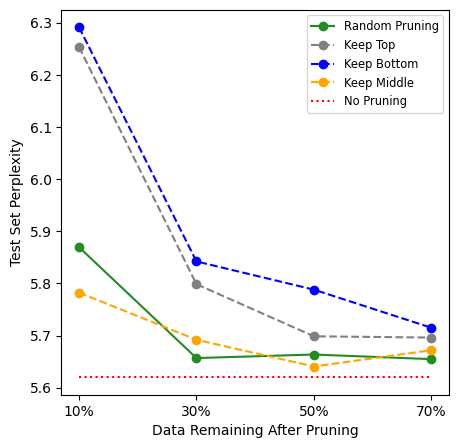}
  \caption{EL2N}
  \label{fig:el2n_only}
\end{subfigure}%
\begin{subfigure}{.5\textwidth}
  \centering
  \includegraphics[width=.7\linewidth]{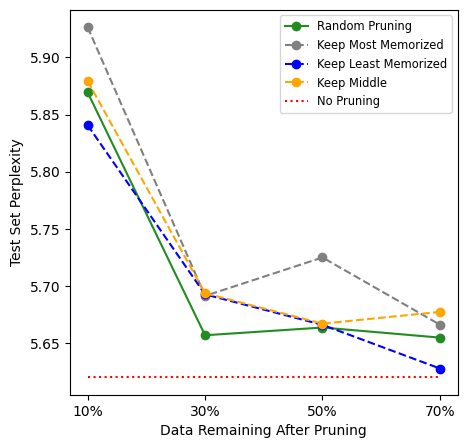}
  \caption{Memorization}
  \label{fig:memorization_only}
\end{subfigure}
\caption{Evaluation of different subset selection criteria for two pruning metrics: (a) EL2N and (b) Memorization.}
\end{figure}

Next, Figure~\ref{fig:memorization_only}(a) shows the results for the \textsc{EL2N} metric.The \texttt{middle} subset is also the best variant for \textsc{EL2N}. While the best performing run does not outperform the baseline, the best performance is achieved when retaining 50\% of the \texttt{middle} subset, outperforming the model trained on 70\% of the dataset, similar to the results when using \textsc{perplexity}. As the \texttt{middle} subset grows, it begins to overlap with the easiest examples, degrading performance. In section \ref{sec:eary_reference_models}, we discuss how different model checkpoints influence the effectiveness of the \textsc{EL2N} metric.

Finally, when using \textsc{memorization factor} as a pruning metric, keeping the least memorized samples (\texttt{bottom} subset) generally performs best. Figure \ref{fig:memorization_only}(b) shows model performances for this metric. We observe that the most competitive variant of the memorization metric is the \texttt{bottom} 70\% of the distribution. Memorization never outperforms the no-pruning baseline.

\subsection{Simple Pruning Metrics Outperform More Sophisticated Approaches}
\label{sec:best_runs}
\begin{figure}[!htb]
    \centering
    \includegraphics[width=.5\textwidth]{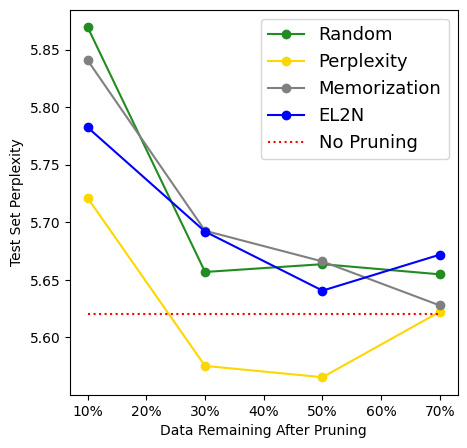}
    \caption{The top performing variants of the different pruning methods, compared across various dataset sizes. Random pruning and no-pruning are included as baselines. Perplexity-based pruning consistently surpasses both alternative metrics and the no pruning experiments. See Section \ref{sec:best_runs} for details on the featured variants.}
    \label{fig:metrics_comp}
\end{figure}

In Figure~\ref{fig:metrics_comp} we present results comparing the performance of the best variant of each pruning metric: (1) retaining the \texttt{middle} of the distribution of \textsc{Perplexity} scores by the fully trained 52B reference model, (2) retaining the \texttt{bottom} of the distribution of the \textsc{Memorization Factor} (least memorized samples), and (3) retaining the \texttt{middle} of the distribution of \textsc{EL2N} scores from the 1000 step checkpoint. 
We also include results for our baselines: a model trained on the entirety of the training data $\mathcal{D}$ and models trained on randomly pruned data.
Our results show that training on the \texttt{middle} subset using \textsc{Perplexity} outperforms other pruning metrics across all dataset sizes. For some variants, it also outperforms training on the entire dataset. For example, at 30\% and 50\% of the original dataset size, \textsc{Perplexity} outperforms the full dataset size. Compared with the no-pruning baseline, pruning to the \texttt{middle} 50\% of the perplexity distribution leads to a 0.97\% improvement in perplexity. Using only the \texttt{middle} 30\% of the data achieves nearly the same performance, with a 0.80\% improvement over the no-pruning baseline. 

Compared with random selection, pruning using \textsc{Perplexity} results in significantly higher model performance than random pruning across all data ratios (Figure~\ref{fig:metrics_comp}). For \textsc{memorization} and \textsc{EL2N} pruning metrics, both achieve similar performances to random pruning despite being far more computationally expensive. 

\subsection{Pruning Benefits from Using Larger Reference Models} 

Given that the most competitive variant perplexity uses a reference model to compute scores, we expect that the size of the reference model will have a significant impact on the data pruned. Figure \ref{fig:impact_of_size} shows the trained model performances after pruning with \textsc{perplexity} calculated with reference models ranging from 124M to 52B parameters. We find that increasing reference model size improves trained model performance over the no-pruning baseline when either the \texttt{middle} or \texttt{top} subsets are used. 
Data pruning using the perplexity scores generated from a 52B parameter reference model achieves a 2.2\% improvement in perplexity over the best-performing trained model from the 124M parameter reference model experiments.
Furthermore, for 13B and 52B reference models, we observe better performances with less training data when keeping the middle and top subsets. For both of these larger models, retaining the \texttt{middle} 30\% and 50\% of the training data produces pruned models that outperform the pruned models trained on the \texttt{middle} 70\% of the training set.

\begin{figure}[t]
\centering
\includegraphics[width=0.5\textwidth]{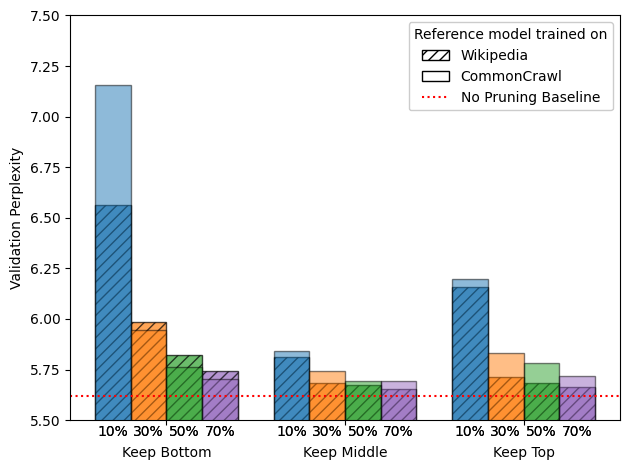}
    \caption{Performance of different pruning strategies using two different reference models: one trained on Wikipedia and one trained on CommonCrawl. A reference model trained on Wikipedia (an example of a clean noise-free corpus) achieves consistently lower validation perplexity compared to a reference model trained on a noisier CommonCrawl in our two robust settings (\texttt{middle} and \texttt{top}).}
    \label{fig:impact_of_source}
\end{figure}

We note that the effects of subset selection, such as the \texttt{bottom} subset performing worse, approximately scale with the size of the reference models. The larger reference models' \texttt{bottom} subset training runs perform even worse than their smaller counterparts when retaining the same percentage of the training set. This overall points to the consistent finding that larger models are better calibrated at computing a useful data pruning ranking.

\subsection{Improved Pruning Signals Result from Reference Models Trained on Cleaner Data}
In this section we ask: \textit{does the data the reference model is trained on impact the quality of the ranking?} We compare the perplexity rankings generated by reference models trained on two different corpora: Wikipedia and CommonCrawl. 
We investigate whether a model trained on Wikipedia, a dataset frequently hand-picked as a high-quality dataset \citep{xie2023data, wenzek-etal-2020-ccnet}, generates more effective pruning signals for perplexity rankings.
In Figure \ref{fig:impact_of_source}, we compare the performance of the two variants across different pruning percentages and subset selections. 
We observe that in the two optimal selection variants from the general reference models (\texttt{middle} and \texttt{top}) a model trained on Wikipedia consistently yields lower validation perplexity compared to a model trained on CommonCrawl. 
Wikipedia's best variant, pruning to the middle 70\%, outperforms CommonCrawl's best variant, also pruning to the middle 70\%, by 0.69\%. This finding overall suggests that investing in a high quality reference model to generate rankings results in more effective data pruning. Reference models trained on higher quality data are better at identifying a subset of data points most conducive to model performance. 

\subsection{Early Reference Model Checkpoints Serve as Effective Scoring Models} 
\label{sec:eary_reference_models}
\begin{figure*}[htb]
    \centering
    \includegraphics[width=0.85\textwidth]{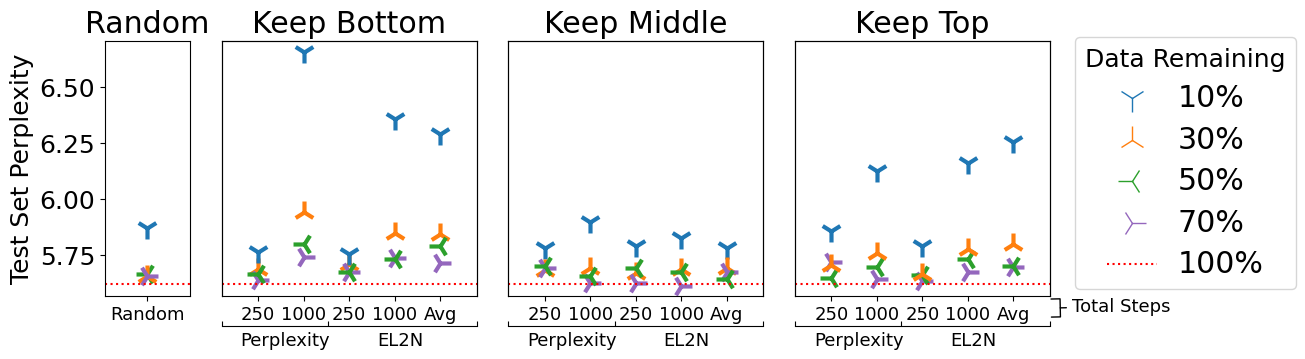}
    \caption{The impact of using an early checkpoint of the reference model in pruning based on Perplexity and EL2N metrics.}
    \label{fig:early_ppl}
\end{figure*}

Motivated by several works that have found that there is a signal in early training checkpoints \citep{paul2023deep, Agarwal_2022_CVPR, siddiqui2022metadata}, we investigate whether early checkpoint of a reference model during training offers adequate signal for calculating discriminative pruning scores.
We study \textsc{perplexity} and \textsc{EL2N} scores obtained from two early checkpoints: after training on approximately 14\% and 55\% of the full training dataset (250 and 1000 training steps respectively).
Figure \ref{fig:early_ppl} showcases the results of these experiments. 
Examining the 14\% checkpoint for both perplexity and EL2N, we notice minimal variance across percentages and subset selection criteria. Performance across subsets changes considerably less than either the 55\% checkpoint or the fully trained models.

Given this, we deduce that training on only 14\% of the data is inadequate for our reference model to offer precise pruning scores.
In contrast, the 55\% reference models perform in a similar manner to the fully trained models, performing best with the \texttt{middle} subset, worst with the \texttt{bottom} subset, and comparably with the \texttt{top} subset. 
Fully training the reference model is shown not to be necessary to uphold comparable performance. Halving the reference model training steps proves effective, enabling the utilization of early checkpoints. In practice, we expect many practitioners to use off the shelf models for computing perplexity and may not need to carry the cost of pretraining a reference model from random initialization. 

We also show performance for EL2N scores averaged across 10 reference models, initialized with different random seeds. We selected the 55\% reference models given our previous result.

While the best pruned models using the averaged EL2N score did not outperform the best pruned models trained on only one reference model's EL2N score, the pattern of performance more similarly mirrors what we see with the larger, fully trained reference models.
Specifically, in the \texttt{middle} subset, using 50\% of the dataset outperforms using 70\%. When constrained to the \texttt{bottom} subset, performance more clearly monotonically degrades when using less data than when using the 55\% reference model, whereas the earlier checkpoint has comparable performance when retaining 30, 50, and 70\% of the data.
This implies that averaging scores across reference models helps hone the pruning signal, identifying subsets ``easy" or ``hard" subsets in more similar ways to larger models. 

\subsection{Perplexity-based Pruning Improvements Generalize to Larger Scale Models}

\begin{figure}[htb]
\centering
    \includegraphics[width=0.5\textwidth]{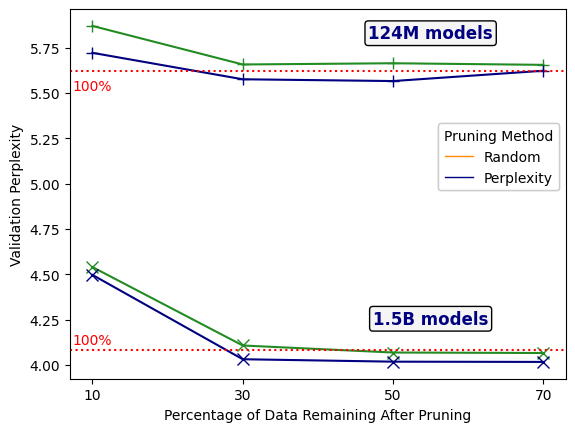}
    \caption{Comparing the best performing pruning method (keeping the \texttt{middle} subset using a 52B parameter reference model) with random pruning at two distinct pruned model scales. The improvement in performance of a perplexity-based pruning approach carries from 124M to 1.5B parameter models.}
    \label{fig:1.5B}
\end{figure}

We take our strongest pruning variant -- \textsc{perplexity} computed using a 52B parameter reference model while retaining the \texttt{middle} subset -- to explore the robustness of our findings at a larger scale by validating our findings on a 1.5B model. Figure~\ref{fig:1.5B} shows pruning scaling from 124M to 1.5B parameter models.
Training a 1.5B model, we observe that random pruning performs considerably well, even reaching levels below the no-pruning run.
Nonetheless, perplexity-based pruning achieves better results than random pruning across all pruning percentages. 
The improvement observed with perplexity-based pruning over random pruning follows a consistent pattern for both the 124M and 1.5B models. This demonstrates the scalability of our approach to a large-scale pretraining setting.

\subsection{Downstream Evaluation on GLUE}
\begin{table*}
\centering
\caption{Mean accuracy and standard deviation of the best variants of each pruning algorithm for GLUE classification tasks. Underlined results surpass the baseline performance with no pruning. The best results for each task are marked in bold. Results are reported for 5 runs of each model, trained for 3 epochs with a learning rate of $1e-5$.}
\label{tab:glue}
\resizebox{1\textwidth}{!}{%
\begin{tabular}{@{}lcccccccc@{}}
\toprule
 & \textbf{Data Remaining} & \textbf{SST2} & \textbf{MRPC} & \textbf{QQP} & \textbf{QNLI} & \textbf{RTE} & \textbf{WNLI} \\ \midrule\midrule
\textbf{No Pruning} & 100\% & 78.15$_{0.002}$ & 64.32$_{0.021}$ & 76.55$_{0.001}$ & 65.40$_{0.006}$ & 49.69$_{0.024}$ & 51.56$_{0.040}$ \\ \midrule
 \multirow{4}{*}{\textbf{\begin{tabular}[c]{@{}l@{}}Random\\ Pruning \end{tabular}}} & 70\% & 77.92$_{0.002}$ & \underline{65.21}$_{0.017}$ & \underline{76.58}$_{0.002}$ & 65.11$_{0.006}$ & 49.69$_{0.013}$ & 48.44$_{0.038}$ \\
 & 50\% & \underline{78.19}$_{0.003}$ & \underline{65.16}$_{0.020}$ & 76.40$_{0.001}$ & \underline{65.44}$_{0.006}$ & \underline{49.92}$_{0.009}$ & 49.69$_{0.062}$ \\
 & 30\% & 77.29$_{0.007}$ & \textbf{66.04}$_{0.017}$ & 76.36$_{0.001}$ & 65.22$_{0.005}$ & \underline{51.33}$_{0.024}$ & 50.31$_{0.057}$ \\
 & 10\% & 76.44$_{0.006}$ & \underline{65.83}$_{0.021}$ & 75.91$_{0.001}$ & 64.40$_{0.007}$ & \underline{50.70}$_{0.007}$ & 50.62$_{0.016}$ \\
 \midrule
\multirow{4}{*}{\textbf{\begin{tabular}[c]{@{}l@{}}Memorization\\ Bottom subset\end{tabular}}} & 70\% & 77.29$_{0.006}$ & \underline{64.38}$_{0.016}$ & 76.42$_{0.001}$ & \underline{66.03}$_{0.007}$ & 49.06$_{0.021}$ & 49.06$_{0.042}$ \\
 & 50\% & 77.89$_{0.006}$ & \underline{65.47}$_{0.017}$ & 76.51$_{0.001}$ & \underline{65.99}$_{0.005}$ & \underline{49.77}$_{0.013}$ & 50.31$_{0.048}$ \\
 & 30\% & \textbf{78.52}$_{0.004}$ & \underline{65.89}$_{0.016}$ & 76.48$_{0.001}$ & \underline{65.91}$_{0.006}$ & \underline{50.31}$_{0.009}$ & \textbf{54.38}$_{0.061}$ \\
 & 10\% & 76.64$_{0.004}$ & \underline{65.16}$_{0.015}$ & 76.11$_{0.001}$ & 64.61$_{0.006}$ & \underline{50.39}$_{0.016}$ & \underline{51.88}$_{0.059}$ \\
\midrule
\multirow{4}{*}{\textbf{\begin{tabular}[c]{@{}l@{}}EL2N\\ Middle subset\end{tabular}}} & 70\% & \underline{78.61}$_{0.008}$ & \underline{66.46}$_{0.018}$ & \underline{76.93}$_{0.001}$ & \underline{67.00}$_{0.005}$ & 48.67$_{0.017}$ & 50.00$_{0.058}$ \\
 & 50\% & \underline{79.17}$_{0.007}$ & \underline{65.42}$_{0.016}$ & 76.35$_{0.001}$ & 62.43$_{0.007}$ & \underline{51.41}$_{0.028}$ & 51.56$_{0.049}$ \\
 & 30\% & \underline{78.98}$_{0.005}$ & \underline{65.41}$_{0.012}$ & \textbf{77.47}$_{0.001}$ & \underline{68.63}$_{0.005}$ & 49.69$_{0.022}$ & \underline{55.31}$_{0.067}$ \\
 & 10\% & \underline{78.31}$_{0.006}$ & 63.38$_{0.016}$ & \underline{76.93}$_{0.001}$ & 65.34$_{0.006}$ & \textbf{51.95}$_{0.021}$ & 51.25$_{0.064}$ \\
 \midrule
\multirow{4}{*}{\textbf{\begin{tabular}[c]{@{}l@{}}Perplexity (52B)\\ Middle subset\end{tabular}}} & 70\% & \underline{78.40}$_{0.004}$ & \underline{64.43}$_{0.020}$ & \underline{76.68}$_{0.001}$ & \textbf{66.74}$_{0.007}$ & \underline{50.16}$_{0.023}$ & 49.06$_{0.012}$ \\
 & 50\% & 78.01$_{0.006}$ & \underline{64.37}$_{0.021}$ & 76.82$_{0.001}$ & \underline{66.00}$_{0.004}$ & \underline{50.62}$_{0.023}$ & 50.31$_{0.021}$ \\
 & 30\% & 77.34$_{0.005}$ & \underline{64.84}$_{0.023}$ & \underline{76.76}$_{0.001}$ & \underline{65.89}$_{0.002}$ & \underline{50.86}$_{0.009}$ & 50.94$_{0.031}$ \\
 & 10\% & 77.66$_{0.006}$ & \underline{65.36}$_{0.017}$ & 76.40$_{0.001}$ & \underline{66.52}$_{0.007}$ & \underline{51.17}$_{0.012}$ & \underline{53.44}$_{0.040}$ \\ \bottomrule
\end{tabular}
}
\end{table*}
Previously, we demonstrated various ways of pruning the pretraining data and training models with different data sizes.
Considering that the pretraining stage primarily focuses on knowledge acquisition~\citep{zhou2023lima}, we inquire about the potential ripple effects of pruning data during pretraining when these models are subsequently finetuned on downstream tasks.
To analyze the impact of different pruning strategies on LLM capabilities, we finetune and evaluate models on a subset of the GLUE tasks~\citep{wang2019glue}.
Results are presented in Table \ref{tab:glue}.
We observe that pruning the pretraining dataset consistently improves performance across all tasks. 
While no single pruning strategy (combining both pruning metric and percentage of remaining data) stands out as superior across all tasks, the absence of a universally dominant approach is consistent with earlier findings in the literature~\citep{gao2021empirical}.
We observe that retaining only 30\% of the least memorized instances yields optimal results for SST2 and WNLI tasks.
With \textsc{perplexity} based pruning, the best performance is obtained on QQP and QNLI tasks by keeping 50\% and 70\% of the training data, respectively.
Even random pruning shows improvements in certain tasks, underscoring the significance of downsampling when handling noisy data during the pretraining stage to mitigate potential learning degradation.

\section{Related Work}
\label{sec:rel_work}

\subsection{Rule-Based Data Pruning in NLP}
Significant portions of web-scraped data used for language model pretraining have been shown to be of low quality, machine-generated spam, pornographic content~\citep{10.1162/tacl_a_00447}. 
Selection processes to determine what should be included in large-scale datasets have centered on rule-based filters and heuristics \citep{bane-etal-2022-comparison}, such as keeping only text written in English~\citep{raffel2020exploring, rae2022scaling} or removing sequences containing blocklisted words~\citep{raffel2020exploring}. There are also quality-based rules such as removing duplicated samples~\citep{zhang2022opt} or filtering sentences that do not fit a certain amount of words \citep{raffel2020exploring, rae2022scaling}. 
Rule-based approaches for data filtering have shown controversial effects on model performance, with some works advertising improvements on language modeling capabilities \citep{penedo2023refinedweb,raffel2020exploring}, while others do not~\citep{black-etal-2022-gpt, biderman2023pythia}. Also, heuristics are prone to undesired outcomes due to their simplicity. For instance ~\citet{dodge2021documenting} show how removing blocklisted words disproportionately removes text from and about minority individuals.

\subsection{Metric-Based Data Pruning in NLP}
Recent work on metric-based pruning has mainly focused on pruning data from the fine-tuning stage of LLMs~\citep{attendu2023nlu, xie2023data} most probably due to the prohibitive cost of pruning at the pretraining scale. \citet{attendu2023nlu} perform dynamic pruning during the fine-tuning stage by establishing a curriculum of samples based on their EL2N scores~\citep{paul2023deep}. Similarly, we benchmark EL2N as a static data-pruning metric for language datasets.
Our work joins the few others that aim to reduce pretraining dataset sizes~\citep{xie2023doremi,chen-2023-large,abbas2023semdedup}. \citet{abbas2023semdedup} apply their deduplication method based on embeddings to further improve the performance of a previously filtered dataset. We also perform pruning on previously filtered datasets, aiming to enhance performance further. Previously, perplexity has been used to filter datasets~\citep{muennighoff2023scaling, wenzek-etal-2020-ccnet, laurençon2023bigscience}, but its pruning capabilities have been underexplored. \citet{laurençon2023bigscience} and \citet{muennighoff2023scaling} filter out high-perplexity samples from their corpus as those are framed as unnatural language and harmful for performance according to their reference domain, which is Wikipedia. In contrast, we benchmark pruning to low perplexity values and high and medium-valued subsets of a dataset's distribution to understand which is the most valuable section for pretraining at scale. We also explore different reference model sizes and training sets.

\subsection{Data pruning in Computer Vision} 
The majority of work to date on data pruning \citep{sorscher2023neural} and isolating data subsets \citep{siddiqui2022metadata, mindermann2022prioritized} using model signal has centered on computer vision. These are typically structured in a supervised setting. In contrast, our focus is on a large-scale NLP pretraining where the objective is unsupervised pretraining. Most relevant to our method is work by \citet{sorscher2023neural} which empirically studies reducing datasets in a teacher/trained regime, using a teacher model's margin as a pruning metric. They find that, with abundant data, training only on the hardest examples yields better performance, while conversely when data is scarce, training on only the easiest example yields better performance. 

\section{Conclusion}
In this study, we thoroughly investigate diverse data pruning methods for pretraining LLMs with billions of parameters and with datasets containing billions of tokens.
We showed that when properly applied, data pruning consistently improves model performance.
We also find that training on the ``easiest" examples in a dataset degrades performance, where ``easiest" is defined as the lowest scoring examples according to a metric based on a reference model. 
Simple methods that rank instances based on their perplexity demonstrate superior performance compared to more elaborate approaches such as memorization.
Models trained on as little as half of the data selected by perplexity achieve up to 1.5\% improvement over models trained on the full dataset.
Additionally, we establish the consistency of our findings as we scale the model sizes.
While scaling up the amount of data LLMs are trained on remains a popular avenue for improving models, our work demonstrates that carefully pruning these large training corpora is also a fruitful direction for making models better.

\bibliography{main}

\begin{thebibliography}{48}
\providecommand{\natexlab}[1]{#1}
\providecommand{\url}[1]{\texttt{#1}}
\expandafter\ifx\csname urlstyle\endcsname\relax
  \providecommand{\doi}[1]{doi: #1}\else
  \providecommand{\doi}{doi: \begingroup \urlstyle{rm}\Url}\fi

\bibitem[Abbas et~al.(2023)Abbas, Tirumala, Simig, Ganguli, and
  Morcos]{abbas2023semdedup}
Amro Abbas, Kushal Tirumala, Dániel Simig, Surya Ganguli, and Ari~S. Morcos.
\newblock Semdedup: Data-efficient learning at web-scale through semantic
  deduplication, 2023.

\bibitem[Agarwal et~al.(2022)Agarwal, D'souza, and Hooker]{Agarwal_2022_CVPR}
Chirag Agarwal, Daniel D'souza, and Sara Hooker.
\newblock Estimating example difficulty using variance of gradients.
\newblock In \emph{Proceedings of the IEEE/CVF Conference on Computer Vision
  and Pattern Recognition (CVPR)}, pp.\  10368--10378, June 2022.

\bibitem[Anil et~al.(2023)Anil, Dai, Firat, Johnson, Lepikhin, Passos, Shakeri,
  Taropa, Bailey, Chen, Chu, Clark, Shafey, Huang, Meier-Hellstern, Mishra,
  Moreira, Omernick, Robinson, Ruder, Tay, Xiao, Xu, Zhang, Abrego, Ahn,
  Austin, Barham, Botha, Bradbury, Brahma, Brooks, Catasta, Cheng, Cherry,
  Choquette-Choo, Chowdhery, Crepy, Dave, Dehghani, Dev, Devlin, Díaz, Du,
  Dyer, Feinberg, Feng, Fienber, Freitag, Garcia, Gehrmann, Gonzalez, Gur-Ari,
  Hand, Hashemi, Hou, Howland, Hu, Hui, Hurwitz, Isard, Ittycheriah, Jagielski,
  Jia, Kenealy, Krikun, Kudugunta, Lan, Lee, Lee, Li, Li, Li, Li, Li, Lim, Lin,
  Liu, Liu, Maggioni, Mahendru, Maynez, Misra, Moussalem, Nado, Nham, Ni,
  Nystrom, Parrish, Pellat, Polacek, Polozov, Pope, Qiao, Reif, Richter, Riley,
  Ros, Roy, Saeta, Samuel, Shelby, Slone, Smilkov, So, Sohn, Tokumine, Valter,
  Vasudevan, Vodrahalli, Wang, Wang, Wang, Wang, Wieting, Wu, Xu, Xu, Xue, Yin,
  Yu, Zhang, Zheng, Zheng, Zhou, Zhou, Petrov, and Wu]{anil2023palm}
Rohan Anil, Andrew~M. Dai, Orhan Firat, Melvin Johnson, Dmitry Lepikhin,
  Alexandre Passos, Siamak Shakeri, Emanuel Taropa, Paige Bailey, Zhifeng Chen,
  Eric Chu, Jonathan~H. Clark, Laurent~El Shafey, Yanping Huang, Kathy
  Meier-Hellstern, Gaurav Mishra, Erica Moreira, Mark Omernick, Kevin Robinson,
  Sebastian Ruder, Yi~Tay, Kefan Xiao, Yuanzhong Xu, Yujing Zhang,
  Gustavo~Hernandez Abrego, Junwhan Ahn, Jacob Austin, Paul Barham, Jan Botha,
  James Bradbury, Siddhartha Brahma, Kevin Brooks, Michele Catasta, Yong Cheng,
  Colin Cherry, Christopher~A. Choquette-Choo, Aakanksha Chowdhery, Clément
  Crepy, Shachi Dave, Mostafa Dehghani, Sunipa Dev, Jacob Devlin, Mark Díaz,
  Nan Du, Ethan Dyer, Vlad Feinberg, Fangxiaoyu Feng, Vlad Fienber, Markus
  Freitag, Xavier Garcia, Sebastian Gehrmann, Lucas Gonzalez, Guy Gur-Ari,
  Steven Hand, Hadi Hashemi, Le~Hou, Joshua Howland, Andrea Hu, Jeffrey Hui,
  Jeremy Hurwitz, Michael Isard, Abe Ittycheriah, Matthew Jagielski, Wenhao
  Jia, Kathleen Kenealy, Maxim Krikun, Sneha Kudugunta, Chang Lan, Katherine
  Lee, Benjamin Lee, Eric Li, Music Li, Wei Li, YaGuang Li, Jian Li, Hyeontaek
  Lim, Hanzhao Lin, Zhongtao Liu, Frederick Liu, Marcello Maggioni, Aroma
  Mahendru, Joshua Maynez, Vedant Misra, Maysam Moussalem, Zachary Nado, John
  Nham, Eric Ni, Andrew Nystrom, Alicia Parrish, Marie Pellat, Martin Polacek,
  Alex Polozov, Reiner Pope, Siyuan Qiao, Emily Reif, Bryan Richter, Parker
  Riley, Alex~Castro Ros, Aurko Roy, Brennan Saeta, Rajkumar Samuel, Renee
  Shelby, Ambrose Slone, Daniel Smilkov, David~R. So, Daniel Sohn, Simon
  Tokumine, Dasha Valter, Vijay Vasudevan, Kiran Vodrahalli, Xuezhi Wang,
  Pidong Wang, Zirui Wang, Tao Wang, John Wieting, Yuhuai Wu, Kelvin Xu, Yunhan
  Xu, Linting Xue, Pengcheng Yin, Jiahui Yu, Qiao Zhang, Steven Zheng,
  Ce~Zheng, Weikang Zhou, Denny Zhou, Slav Petrov, and Yonghui Wu.
\newblock Palm 2 technical report, 2023.

\bibitem[Attendu \& Corbeil(2023)Attendu and Corbeil]{attendu2023nlu}
Jean-Michel Attendu and Jean-Philippe Corbeil.
\newblock Nlu on data diets: Dynamic data subset selection for nlp
  classification tasks, 2023.

\bibitem[Bane et~al.(2022)Bane, Uguet, Stribi{\.z}ew, and
  Zaretskaya]{bane-etal-2022-comparison}
Fred Bane, Celia~Soler Uguet, Wiktor Stribi{\.z}ew, and Anna Zaretskaya.
\newblock A comparison of data filtering methods for neural machine
  translation.
\newblock In \emph{Proceedings of the 15th Biennial Conference of the
  Association for Machine Translation in the Americas (Volume 2: Users and
  Providers Track and Government Track)}, pp.\  313--325, Orlando, USA,
  September 2022. Association for Machine Translation in the Americas.
\newblock URL \url{https://aclanthology.org/2022.amta-upg.22}.

\bibitem[Biderman et~al.(2023{\natexlab{a}})Biderman, Prashanth, Sutawika,
  Schoelkopf, Anthony, Purohit, and Raff]{biderman2023emergent}
Stella Biderman, USVSN~Sai Prashanth, Lintang Sutawika, Hailey Schoelkopf,
  Quentin Anthony, Shivanshu Purohit, and Edward Raff.
\newblock Emergent and predictable memorization in large language models,
  2023{\natexlab{a}}.

\bibitem[Biderman et~al.(2023{\natexlab{b}})Biderman, Schoelkopf, Anthony,
  Bradley, O'Brien, Hallahan, Khan, Purohit, Prashanth, Raff, Skowron,
  Sutawika, and van~der Wal]{biderman2023pythia}
Stella Biderman, Hailey Schoelkopf, Quentin Anthony, Herbie Bradley, Kyle
  O'Brien, Eric Hallahan, Mohammad~Aflah Khan, Shivanshu Purohit, USVSN~Sai
  Prashanth, Edward Raff, Aviya Skowron, Lintang Sutawika, and Oskar van~der
  Wal.
\newblock Pythia: A suite for analyzing large language models across training
  and scaling, 2023{\natexlab{b}}.

\bibitem[Black et~al.(2022)Black, Biderman, Hallahan, Anthony, Gao, Golding,
  He, Leahy, McDonell, Phang, Pieler, Prashanth, Purohit, Reynolds, Tow, Wang,
  and Weinbach]{black-etal-2022-gpt}
Sidney Black, Stella Biderman, Eric Hallahan, Quentin Anthony, Leo Gao,
  Laurence Golding, Horace He, Connor Leahy, Kyle McDonell, Jason Phang,
  Michael Pieler, Usvsn~Sai Prashanth, Shivanshu Purohit, Laria Reynolds,
  Jonathan Tow, Ben Wang, and Samuel Weinbach.
\newblock {GPT}-{N}eo{X}-20{B}: An open-source autoregressive language model.
\newblock In \emph{Proceedings of BigScience Episode {\#}5 -- Workshop on
  Challenges {\&} Perspectives in Creating Large Language Models}, pp.\
  95--136, virtual+Dublin, May 2022. Association for Computational Linguistics.
\newblock \doi{10.18653/v1/2022.bigscience-1.9}.
\newblock URL \url{https://aclanthology.org/2022.bigscience-1.9}.

\bibitem[Brown et~al.(2020)Brown, Mann, Ryder, Subbiah, Kaplan, Dhariwal,
  Neelakantan, Shyam, Sastry, Askell, Agarwal, Herbert-Voss, Krueger, Henighan,
  Child, Ramesh, Ziegler, Wu, Winter, Hesse, Chen, Sigler, Litwin, Gray, Chess,
  Clark, Berner, McCandlish, Radford, Sutskever, and Amodei]{brown2020language}
Tom~B. Brown, Benjamin Mann, Nick Ryder, Melanie Subbiah, Jared Kaplan,
  Prafulla Dhariwal, Arvind Neelakantan, Pranav Shyam, Girish Sastry, Amanda
  Askell, Sandhini Agarwal, Ariel Herbert-Voss, Gretchen Krueger, Tom Henighan,
  Rewon Child, Aditya Ramesh, Daniel~M. Ziegler, Jeffrey Wu, Clemens Winter,
  Christopher Hesse, Mark Chen, Eric Sigler, Mateusz Litwin, Scott Gray,
  Benjamin Chess, Jack Clark, Christopher Berner, Sam McCandlish, Alec Radford,
  Ilya Sutskever, and Dario Amodei.
\newblock Language models are few-shot learners, 2020.

\bibitem[Cao et~al.(2023)Cao, Kang, and Sun]{cao2023instruction}
Yihan Cao, Yanbin Kang, and Lichao Sun.
\newblock Instruction mining: High-quality instruction data selection for large
  language models, 2023.

\bibitem[Carlini et~al.(2021)Carlini, Tramer, Wallace, Jagielski, Herbert-Voss,
  Lee, Roberts, Brown, Song, Erlingsson, Oprea, and
  Raffel]{carlini2021extracting}
Nicholas Carlini, Florian Tramer, Eric Wallace, Matthew Jagielski, Ariel
  Herbert-Voss, Katherine Lee, Adam Roberts, Tom Brown, Dawn Song, Ulfar
  Erlingsson, Alina Oprea, and Colin Raffel.
\newblock Extracting training data from large language models, 2021.

\bibitem[Carlini et~al.(2023)Carlini, Ippolito, Jagielski, Lee, Tramer, and
  Zhang]{carlini2023quantifying}
Nicholas Carlini, Daphne Ippolito, Matthew Jagielski, Katherine Lee, Florian
  Tramer, and Chiyuan Zhang.
\newblock Quantifying memorization across neural language models, 2023.

\bibitem[Chen(2023)]{chen-2023-large}
Wenhu Chen.
\newblock Large language models are few(1)-shot table reasoners.
\newblock In \emph{Findings of the Association for Computational Linguistics:
  EACL 2023}, pp.\  1120--1130, Dubrovnik, Croatia, May 2023. Association for
  Computational Linguistics.
\newblock URL \url{https://aclanthology.org/2023.findings-eacl.83}.

\bibitem[Dodge et~al.(2021)Dodge, Sap, Marasović, Agnew, Ilharco, Groeneveld,
  Mitchell, and Gardner]{dodge2021documenting}
Jesse Dodge, Maarten Sap, Ana Marasović, William Agnew, Gabriel Ilharco, Dirk
  Groeneveld, Margaret Mitchell, and Matt Gardner.
\newblock Documenting large webtext corpora: A case study on the colossal clean
  crawled corpus, 2021.

\bibitem[Fayyaz et~al.(2022)Fayyaz, Aghazadeh, Modarressi, Pilehvar,
  Yaghoobzadeh, and Kahou]{fayyaz2022bert}
Mohsen Fayyaz, Ehsan Aghazadeh, Ali Modarressi, Mohammad~Taher Pilehvar,
  Yadollah Yaghoobzadeh, and Samira~Ebrahimi Kahou.
\newblock Bert on a data diet: Finding important examples by gradient-based
  pruning, 2022.

\bibitem[Gao(2021)]{gao2021empirical}
Leo Gao.
\newblock An empirical exploration in quality filtering of text data, 2021.

\bibitem[Gao et~al.(2021)Gao, Biderman, Black, Golding, Hoppe, Foster, Phang,
  He, Thite, Nabeshima, Presser, and Leahy]{DBLP:journals/corr/abs-2101-00027}
Leo Gao, Stella Biderman, Sid Black, Laurence Golding, Travis Hoppe, Charles
  Foster, Jason Phang, Horace He, Anish Thite, Noa Nabeshima, Shawn Presser,
  and Connor Leahy.
\newblock The pile: An 800gb dataset of diverse text for language modeling.
\newblock \emph{CoRR}, abs/2101.00027, 2021.
\newblock URL \url{https://arxiv.org/abs/2101.00027}.

\bibitem[He et~al.(2023)He, Yang, Huang, and Zhao]{he2023largescale}
Muyang He, Shuo Yang, Tiejun Huang, and Bo~Zhao.
\newblock Large-scale dataset pruning with dynamic uncertainty, 2023.

\bibitem[Hernandez et~al.(2022)Hernandez, Brown, Conerly, DasSarma, Drain,
  El-Showk, Elhage, Hatfield-Dodds, Henighan, Hume, Johnston, Mann, Olah,
  Olsson, Amodei, Joseph, Kaplan, and McCandlish]{hernandez2022scaling}
Danny Hernandez, Tom Brown, Tom Conerly, Nova DasSarma, Dawn Drain, Sheer
  El-Showk, Nelson Elhage, Zac Hatfield-Dodds, Tom Henighan, Tristan Hume,
  Scott Johnston, Ben Mann, Chris Olah, Catherine Olsson, Dario Amodei,
  Nicholas Joseph, Jared Kaplan, and Sam McCandlish.
\newblock Scaling laws and interpretability of learning from repeated data,
  2022.

\bibitem[Kaplan et~al.(2020)Kaplan, McCandlish, Henighan, Brown, Chess, Child,
  Gray, Radford, Wu, and Amodei]{kaplan2020scaling}
Jared Kaplan, Sam McCandlish, Tom Henighan, Tom~B. Brown, Benjamin Chess, Rewon
  Child, Scott Gray, Alec Radford, Jeffrey Wu, and Dario Amodei.
\newblock Scaling laws for neural language models, 2020.

\bibitem[Kocetkov et~al.(2022)Kocetkov, Li, Allal, Li, Mou, Ferrandis, Jernite,
  Mitchell, Hughes, Wolf, Bahdanau, von Werra, and de~Vries]{kocetkov2022stack}
Denis Kocetkov, Raymond Li, Loubna~Ben Allal, Jia Li, Chenghao Mou,
  Carlos~Muñoz Ferrandis, Yacine Jernite, Margaret Mitchell, Sean Hughes,
  Thomas Wolf, Dzmitry Bahdanau, Leandro von Werra, and Harm de~Vries.
\newblock The stack: 3 tb of permissively licensed source code, 2022.

\bibitem[Kreutzer et~al.(2022)Kreutzer, Caswell, Wang, Wahab, van Esch,
  Ulzii-Orshikh, Tapo, Subramani, Sokolov, Sikasote, Setyawan, Sarin, Samb,
  Sagot, Rivera, Rios, Papadimitriou, Osei, Suarez, Orife, Ogueji, Rubungo,
  Nguyen, Müller, Müller, Muhammad, Muhammad, Mnyakeni, Mirzakhalov,
  Matangira, Leong, Lawson, Kudugunta, Jernite, Jenny, Firat, Dossou, Dlamini,
  de~Silva, Çabuk Ballı, Biderman, Battisti, Baruwa, Bapna, Baljekar, Azime,
  Awokoya, Ataman, Ahia, Ahia, Agrawal, and Adeyemi]{10.1162/tacl_a_00447}
Julia Kreutzer, Isaac Caswell, Lisa Wang, Ahsan Wahab, Daan van Esch,
  Nasanbayar Ulzii-Orshikh, Allahsera Tapo, Nishant Subramani, Artem Sokolov,
  Claytone Sikasote, Monang Setyawan, Supheakmungkol Sarin, Sokhar Samb,
  Benoît Sagot, Clara Rivera, Annette Rios, Isabel Papadimitriou, Salomey
  Osei, Pedro~Ortiz Suarez, Iroro Orife, Kelechi Ogueji, Andre~Niyongabo
  Rubungo, Toan~Q. Nguyen, Mathias Müller, André Müller, Shamsuddeen~Hassan
  Muhammad, Nanda Muhammad, Ayanda Mnyakeni, Jamshidbek Mirzakhalov,
  Tapiwanashe Matangira, Colin Leong, Nze Lawson, Sneha Kudugunta, Yacine
  Jernite, Mathias Jenny, Orhan Firat, Bonaventure F.~P. Dossou, Sakhile
  Dlamini, Nisansa de~Silva, Sakine Çabuk Ballı, Stella Biderman, Alessia
  Battisti, Ahmed Baruwa, Ankur Bapna, Pallavi Baljekar, Israel~Abebe Azime,
  Ayodele Awokoya, Duygu Ataman, Orevaoghene Ahia, Oghenefego Ahia, Sweta
  Agrawal, and Mofetoluwa Adeyemi.
\newblock {Quality at a Glance: An Audit of Web-Crawled Multilingual Datasets}.
\newblock \emph{Transactions of the Association for Computational Linguistics},
  10:\penalty0 50--72, 01 2022.
\newblock ISSN 2307-387X.
\newblock \doi{10.1162/tacl_a_00447}.
\newblock URL \url{https://doi.org/10.1162/tacl\_a\_00447}.

\bibitem[Laurençon et~al.(2023)Laurençon, Saulnier, Wang, Akiki, del Moral,
  Scao, Werra, Mou, Ponferrada, Nguyen, Frohberg, Šaško, Lhoest,
  McMillan-Major, Dupont, Biderman, Rogers, allal, Toni, Pistilli, Nguyen,
  Nikpoor, Masoud, Colombo, de~la Rosa, Villegas, Thrush, Longpre, Nagel,
  Weber, Muñoz, Zhu, Strien, Alyafeai, Almubarak, Vu, Gonzalez-Dios, Soroa,
  Lo, Dey, Suarez, Gokaslan, Bose, Adelani, Phan, Tran, Yu, Pai, Chim, Lepercq,
  Ilic, Mitchell, Luccioni, and Jernite]{laurençon2023bigscience}
Hugo Laurençon, Lucile Saulnier, Thomas Wang, Christopher Akiki,
  Albert~Villanova del Moral, Teven~Le Scao, Leandro~Von Werra, Chenghao Mou,
  Eduardo~González Ponferrada, Huu Nguyen, Jörg Frohberg, Mario Šaško,
  Quentin Lhoest, Angelina McMillan-Major, Gerard Dupont, Stella Biderman, Anna
  Rogers, Loubna~Ben allal, Francesco~De Toni, Giada Pistilli, Olivier Nguyen,
  Somaieh Nikpoor, Maraim Masoud, Pierre Colombo, Javier de~la Rosa, Paulo
  Villegas, Tristan Thrush, Shayne Longpre, Sebastian Nagel, Leon Weber, Manuel
  Muñoz, Jian Zhu, Daniel~Van Strien, Zaid Alyafeai, Khalid Almubarak,
  Minh~Chien Vu, Itziar Gonzalez-Dios, Aitor Soroa, Kyle Lo, Manan Dey,
  Pedro~Ortiz Suarez, Aaron Gokaslan, Shamik Bose, David Adelani, Long Phan,
  Hieu Tran, Ian Yu, Suhas Pai, Jenny Chim, Violette Lepercq, Suzana Ilic,
  Margaret Mitchell, Sasha~Alexandra Luccioni, and Yacine Jernite.
\newblock The bigscience roots corpus: A 1.6tb composite multilingual dataset,
  2023.

\bibitem[Loshchilov \& Hutter(2019)Loshchilov and
  Hutter]{loshchilov2018decoupled}
Ilya Loshchilov and Frank Hutter.
\newblock Decoupled weight decay regularization.
\newblock In \emph{International Conference on Learning Representations}, 2019.
\newblock URL \url{https://openreview.net/forum?id=Bkg6RiCqY7}.

\bibitem[Luccioni \& Viviano(2021)Luccioni and
  Viviano]{luccioni-viviano-2021-whats}
Alexandra Luccioni and Joseph Viviano.
\newblock What{'}s in the box? an analysis of undesirable content in the
  {C}ommon {C}rawl corpus.
\newblock In \emph{Proceedings of the 59th Annual Meeting of the Association
  for Computational Linguistics and the 11th International Joint Conference on
  Natural Language Processing (Volume 2: Short Papers)}, pp.\  182--189,
  Online, August 2021. Association for Computational Linguistics.
\newblock \doi{10.18653/v1/2021.acl-short.24}.
\newblock URL \url{https://aclanthology.org/2021.acl-short.24}.

\bibitem[Mindermann et~al.(2022)Mindermann, Brauner, Razzak, Sharma, Kirsch,
  Xu, Höltgen, Gomez, Morisot, Farquhar, and Gal]{mindermann2022prioritized}
Sören Mindermann, Jan Brauner, Muhammed Razzak, Mrinank Sharma, Andreas
  Kirsch, Winnie Xu, Benedikt Höltgen, Aidan~N. Gomez, Adrien Morisot,
  Sebastian Farquhar, and Yarin Gal.
\newblock Prioritized training on points that are learnable, worth learning,
  and not yet learnt, 2022.

\bibitem[Mitchell et~al.(2023)Mitchell, Luccioni, Lambert, Gerchick,
  McMillan-Major, Ozoani, Rajani, Thrush, Jernite, and
  Kiela]{mitchell2023measuring}
Margaret Mitchell, Alexandra~Sasha Luccioni, Nathan Lambert, Marissa Gerchick,
  Angelina McMillan-Major, Ezinwanne Ozoani, Nazneen Rajani, Tristan Thrush,
  Yacine Jernite, and Douwe Kiela.
\newblock Measuring data, 2023.

\bibitem[Muennighoff et~al.(2023)Muennighoff, Rush, Barak, Scao, Piktus, Tazi,
  Pyysalo, Wolf, and Raffel]{muennighoff2023scaling}
Niklas Muennighoff, Alexander~M. Rush, Boaz Barak, Teven~Le Scao, Aleksandra
  Piktus, Nouamane Tazi, Sampo Pyysalo, Thomas Wolf, and Colin Raffel.
\newblock Scaling data-constrained language models, 2023.

\bibitem[Park et~al.(2022)Park, Papailiopoulos, and Lee]{park2022active}
Dongmin Park, Dimitris Papailiopoulos, and Kangwook Lee.
\newblock Active learning is a strong baseline for data subset selection.
\newblock In \emph{Has it Trained Yet? NeurIPS 2022 Workshop}, 2022.
\newblock URL \url{https://openreview.net/forum?id=PAgpyQ5rGS}.

\bibitem[Paul et~al.(2023)Paul, Ganguli, and Dziugaite]{paul2023deep}
Mansheej Paul, Surya Ganguli, and Gintare~Karolina Dziugaite.
\newblock Deep learning on a data diet: Finding important examples early in
  training, 2023.

\bibitem[Penedo et~al.(2023)Penedo, Malartic, Hesslow, Cojocaru, Cappelli,
  Alobeidli, Pannier, Almazrouei, and Launay]{penedo2023refinedweb}
Guilherme Penedo, Quentin Malartic, Daniel Hesslow, Ruxandra Cojocaru,
  Alessandro Cappelli, Hamza Alobeidli, Baptiste Pannier, Ebtesam Almazrouei,
  and Julien Launay.
\newblock The refinedweb dataset for falcon llm: Outperforming curated corpora
  with web data, and web data only, 2023.

\bibitem[Qin et~al.(2023)Qin, Wang, Zheng, Gu, Peng, Zhou, and
  You]{qin2023infobatch}
Ziheng Qin, Kai Wang, Zangwei Zheng, Jianyang Gu, Xiangyu Peng, Daquan Zhou,
  and Yang You.
\newblock Infobatch: Lossless training speed up by unbiased dynamic data
  pruning, 2023.

\bibitem[Radford et~al.(2018)Radford, Narasimhan, Salimans, Sutskever,
  et~al.]{radford2018improving}
Alec Radford, Karthik Narasimhan, Tim Salimans, Ilya Sutskever, et~al.
\newblock Improving language understanding by generative pre-training.
\newblock 2018.

\bibitem[Rae et~al.(2022)Rae, Borgeaud, Cai, Millican, Hoffmann, Song,
  Aslanides, Henderson, Ring, Young, Rutherford, Hennigan, Menick, Cassirer,
  Powell, van~den Driessche, Hendricks, Rauh, Huang, Glaese, Welbl, Dathathri,
  Huang, Uesato, Mellor, Higgins, Creswell, McAleese, Wu, Elsen, Jayakumar,
  Buchatskaya, Budden, Sutherland, Simonyan, Paganini, Sifre, Martens, Li,
  Kuncoro, Nematzadeh, Gribovskaya, Donato, Lazaridou, Mensch, Lespiau,
  Tsimpoukelli, Grigorev, Fritz, Sottiaux, Pajarskas, Pohlen, Gong, Toyama,
  de~Masson~d'Autume, Li, Terzi, Mikulik, Babuschkin, Clark, de~Las~Casas, Guy,
  Jones, Bradbury, Johnson, Hechtman, Weidinger, Gabriel, Isaac, Lockhart,
  Osindero, Rimell, Dyer, Vinyals, Ayoub, Stanway, Bennett, Hassabis,
  Kavukcuoglu, and Irving]{rae2022scaling}
Jack~W. Rae, Sebastian Borgeaud, Trevor Cai, Katie Millican, Jordan Hoffmann,
  Francis Song, John Aslanides, Sarah Henderson, Roman Ring, Susannah Young,
  Eliza Rutherford, Tom Hennigan, Jacob Menick, Albin Cassirer, Richard Powell,
  George van~den Driessche, Lisa~Anne Hendricks, Maribeth Rauh, Po-Sen Huang,
  Amelia Glaese, Johannes Welbl, Sumanth Dathathri, Saffron Huang, Jonathan
  Uesato, John Mellor, Irina Higgins, Antonia Creswell, Nat McAleese, Amy Wu,
  Erich Elsen, Siddhant Jayakumar, Elena Buchatskaya, David Budden, Esme
  Sutherland, Karen Simonyan, Michela Paganini, Laurent Sifre, Lena Martens,
  Xiang~Lorraine Li, Adhiguna Kuncoro, Aida Nematzadeh, Elena Gribovskaya,
  Domenic Donato, Angeliki Lazaridou, Arthur Mensch, Jean-Baptiste Lespiau,
  Maria Tsimpoukelli, Nikolai Grigorev, Doug Fritz, Thibault Sottiaux, Mantas
  Pajarskas, Toby Pohlen, Zhitao Gong, Daniel Toyama, Cyprien
  de~Masson~d'Autume, Yujia Li, Tayfun Terzi, Vladimir Mikulik, Igor
  Babuschkin, Aidan Clark, Diego de~Las~Casas, Aurelia Guy, Chris Jones, James
  Bradbury, Matthew Johnson, Blake Hechtman, Laura Weidinger, Iason Gabriel,
  William Isaac, Ed~Lockhart, Simon Osindero, Laura Rimell, Chris Dyer, Oriol
  Vinyals, Kareem Ayoub, Jeff Stanway, Lorrayne Bennett, Demis Hassabis, Koray
  Kavukcuoglu, and Geoffrey Irving.
\newblock Scaling language models: Methods, analysis and insights from training
  gopher, 2022.

\bibitem[Raffel et~al.(2020)Raffel, Shazeer, Roberts, Lee, Narang, Matena,
  Zhou, Li, and Liu]{raffel2020exploring}
Colin Raffel, Noam Shazeer, Adam Roberts, Katherine Lee, Sharan Narang, Michael
  Matena, Yanqi Zhou, Wei Li, and Peter~J. Liu.
\newblock Exploring the limits of transfer learning with a unified text-to-text
  transformer, 2020.

\bibitem[Raju et~al.(2021)Raju, Daruwalla, and Lipasti]{raju2021accelerating}
Ravi~S Raju, Kyle Daruwalla, and Mikko Lipasti.
\newblock Accelerating deep learning with dynamic data pruning, 2021.

\bibitem[Sennrich et~al.(2016)Sennrich, Haddow, and Birch]{sennrich2016neural}
Rico Sennrich, Barry Haddow, and Alexandra Birch.
\newblock Neural machine translation of rare words with subword units, 2016.

\bibitem[Siddiqui et~al.(2022)Siddiqui, Rajkumar, Maharaj, Krueger, and
  Hooker]{siddiqui2022metadata}
Shoaib~Ahmed Siddiqui, Nitarshan Rajkumar, Tegan Maharaj, David Krueger, and
  Sara Hooker.
\newblock Metadata archaeology: Unearthing data subsets by leveraging training
  dynamics, 2022.

\bibitem[Sorscher et~al.(2023)Sorscher, Geirhos, Shekhar, Ganguli, and
  Morcos]{sorscher2023neural}
Ben Sorscher, Robert Geirhos, Shashank Shekhar, Surya Ganguli, and Ari~S.
  Morcos.
\newblock Beyond neural scaling laws: beating power law scaling via data
  pruning, 2023.

\bibitem[Taylor et~al.(2022)Taylor, Kardas, Cucurull, Scialom, Hartshorn,
  Saravia, Poulton, Kerkez, and Stojnic]{taylor2022galactica}
Ross Taylor, Marcin Kardas, Guillem Cucurull, Thomas Scialom, Anthony
  Hartshorn, Elvis Saravia, Andrew Poulton, Viktor Kerkez, and Robert Stojnic.
\newblock Galactica: A large language model for science, 2022.

\bibitem[Touvron et~al.(2023)Touvron, Lavril, Izacard, Martinet, Lachaux,
  Lacroix, Rozière, Goyal, Hambro, Azhar, Rodriguez, Joulin, Grave, and
  Lample]{touvron2023llama}
Hugo Touvron, Thibaut Lavril, Gautier Izacard, Xavier Martinet, Marie-Anne
  Lachaux, Timothée Lacroix, Baptiste Rozière, Naman Goyal, Eric Hambro,
  Faisal Azhar, Aurelien Rodriguez, Armand Joulin, Edouard Grave, and Guillaume
  Lample.
\newblock Llama: Open and efficient foundation language models, 2023.

\bibitem[Vaswani et~al.(2023)Vaswani, Shazeer, Parmar, Uszkoreit, Jones, Gomez,
  Kaiser, and Polosukhin]{vaswani2023attention}
Ashish Vaswani, Noam Shazeer, Niki Parmar, Jakob Uszkoreit, Llion Jones,
  Aidan~N. Gomez, Lukasz Kaiser, and Illia Polosukhin.
\newblock Attention is all you need, 2023.

\bibitem[Wang et~al.(2019)Wang, Singh, Michael, Hill, Levy, and
  Bowman]{wang2019glue}
Alex Wang, Amanpreet Singh, Julian Michael, Felix Hill, Omer Levy, and
  Samuel~R. Bowman.
\newblock Glue: A multi-task benchmark and analysis platform for natural
  language understanding, 2019.

\bibitem[Wenzek et~al.(2020)Wenzek, Lachaux, Conneau, Chaudhary, Guzm{\'a}n,
  Joulin, and Grave]{wenzek-etal-2020-ccnet}
Guillaume Wenzek, Marie-Anne Lachaux, Alexis Conneau, Vishrav Chaudhary,
  Francisco Guzm{\'a}n, Armand Joulin, and Edouard Grave.
\newblock {CCN}et: Extracting high quality monolingual datasets from web crawl
  data.
\newblock In \emph{Proceedings of the Twelfth Language Resources and Evaluation
  Conference}, pp.\  4003--4012, Marseille, France, May 2020. European Language
  Resources Association.
\newblock ISBN 979-10-95546-34-4.
\newblock URL \url{https://aclanthology.org/2020.lrec-1.494}.

\bibitem[Xie et~al.(2023{\natexlab{a}})Xie, Pham, Dong, Du, Liu, Lu, Liang, Le,
  Ma, and Yu]{xie2023doremi}
Sang~Michael Xie, Hieu Pham, Xuanyi Dong, Nan Du, Hanxiao Liu, Yifeng Lu, Percy
  Liang, Quoc~V Le, Tengyu Ma, and Adams~Wei Yu.
\newblock Doremi: Optimizing data mixtures speeds up language model
  pretraining.
\newblock \emph{arXiv preprint arXiv:2305.10429}, 2023{\natexlab{a}}.

\bibitem[Xie et~al.(2023{\natexlab{b}})Xie, Santurkar, Ma, and
  Liang]{xie2023data}
Sang~Michael Xie, Shibani Santurkar, Tengyu Ma, and Percy Liang.
\newblock Data selection for language models via importance resampling,
  2023{\natexlab{b}}.

\bibitem[Zhang et~al.(2022)Zhang, Roller, Goyal, Artetxe, Chen, Chen, Dewan,
  Diab, Li, Lin, Mihaylov, Ott, Shleifer, Shuster, Simig, Koura, Sridhar, Wang,
  and Zettlemoyer]{zhang2022opt}
Susan Zhang, Stephen Roller, Naman Goyal, Mikel Artetxe, Moya Chen, Shuohui
  Chen, Christopher Dewan, Mona Diab, Xian Li, Xi~Victoria Lin, Todor Mihaylov,
  Myle Ott, Sam Shleifer, Kurt Shuster, Daniel Simig, Punit~Singh Koura, Anjali
  Sridhar, Tianlu Wang, and Luke Zettlemoyer.
\newblock Opt: Open pre-trained transformer language models, 2022.

\bibitem[Zhou et~al.(2023)Zhou, Liu, Xu, Iyer, Sun, Mao, Ma, Efrat, Yu, Yu,
  Zhang, Ghosh, Lewis, Zettlemoyer, and Levy]{zhou2023lima}
Chunting Zhou, Pengfei Liu, Puxin Xu, Srini Iyer, Jiao Sun, Yuning Mao, Xuezhe
  Ma, Avia Efrat, Ping Yu, Lili Yu, Susan Zhang, Gargi Ghosh, Mike Lewis, Luke
  Zettlemoyer, and Omer Levy.
\newblock Lima: Less is more for alignment, 2023.

\end{thebibliography}

\appendix
\section{Metric Distributions}\label{sec:appendix_distributions}
\begin{figure}[!htb]
 \centering 
 \begin{subfigure}[t]{\linewidth}
 \includegraphics[width=1.0\textwidth]{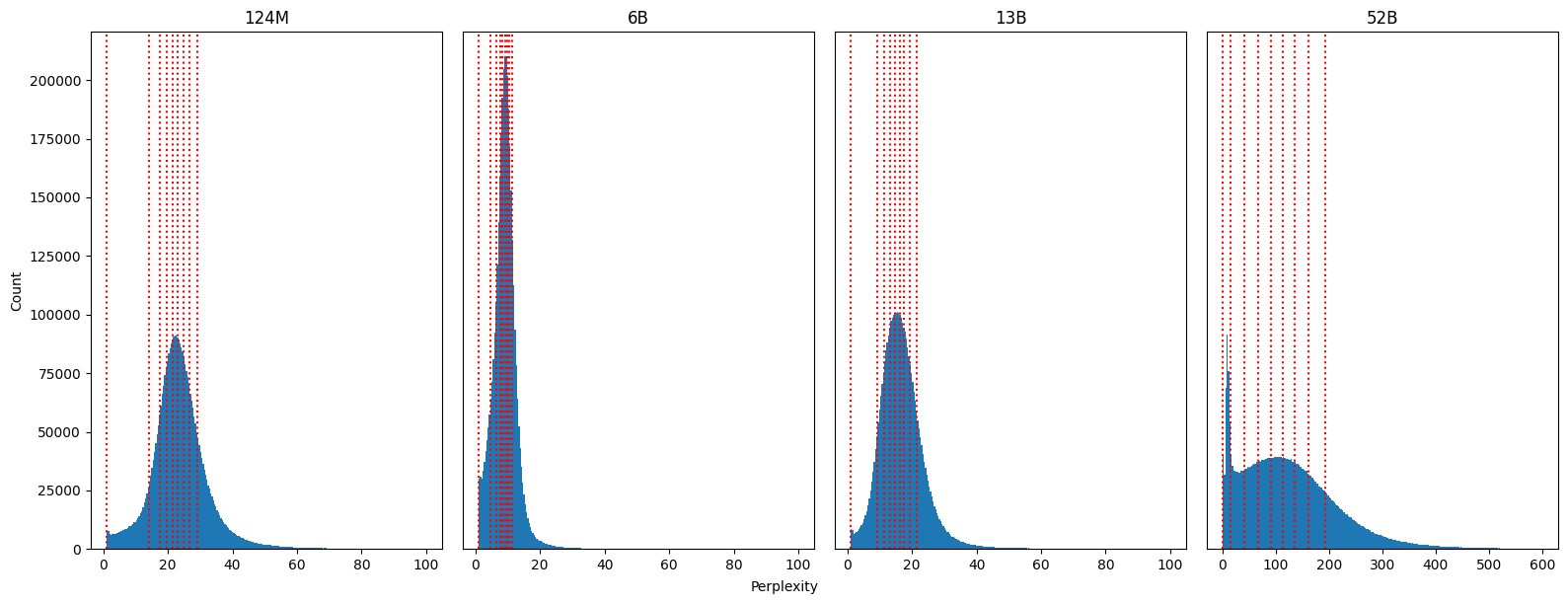}
    \caption{Distributions of Perplexity from different reference models. The dotted lines are placed at each 10th percentile. Please note the differences in axes between graphs. Fewer than .1\% of examples on the extreme high end have been truncate to better display the overall distribution}
    \label{fig:full_distributions}
 \end{subfigure}
 \begin{subfigure}[t]{.49\linewidth}
  \centering
  \includegraphics[width=\linewidth]{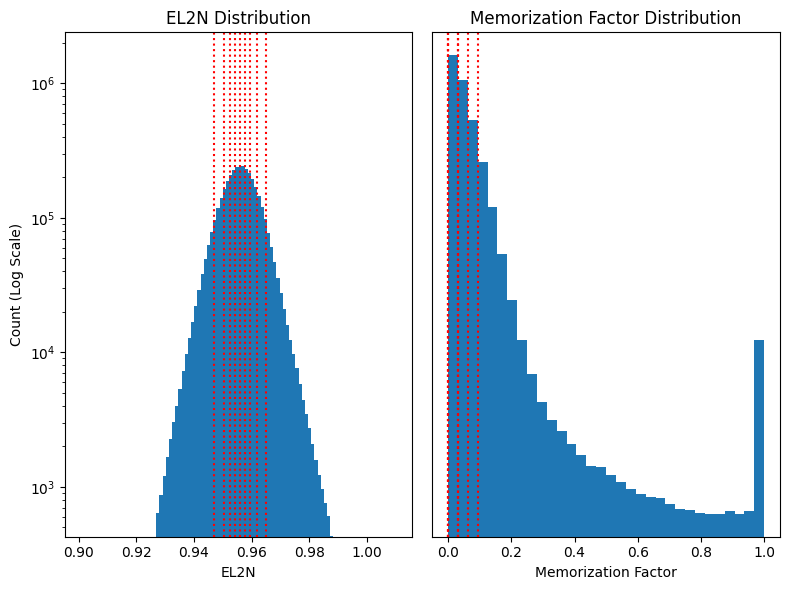}
  \caption{Distributions of the EL2N and Memorization Factor metrics. The dotted lines are placed at each 10th percentile and omitted from Memorization Factor due to overlap. Please note the log-scaled y-axis.}
    \label{fig:metric_distributions}
 \end{subfigure}
 \begin{subfigure}[t]{.49\linewidth}
  \centering
  \includegraphics[width=\linewidth]{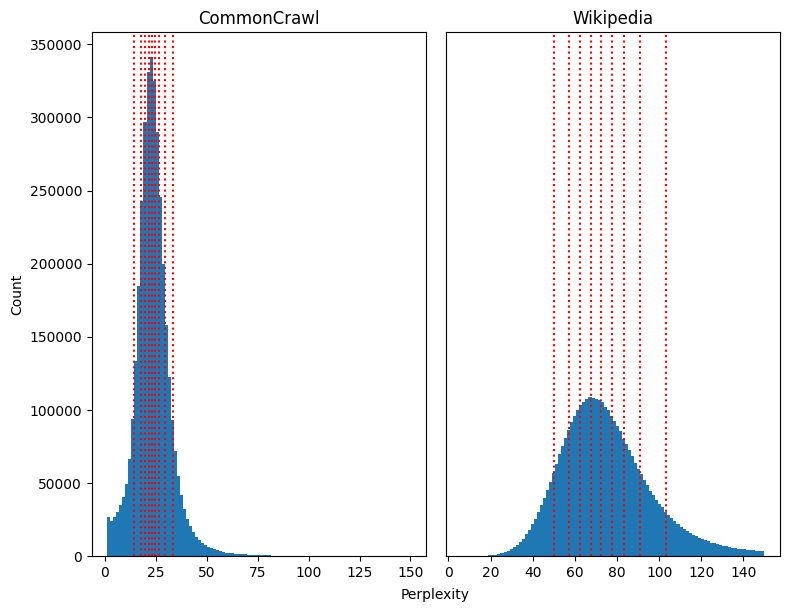}
    \caption{Distributions of Perplexity from reference models trained on Wikipedia and CommonCrawl. The CommonCrawl model is the same as the 124M parameter model in Figure \ref{fig:full_distributions}. The dotted lines are placed at each 10th percentile.}
    \label{fig:source_distributions}
 \end{subfigure}
   \caption{Distributions of different pruning metrics and reference models. }
   \label{fig:all_metric_distributions}
\end{figure}

We present the total distributions of the pruning metrics used in our analysis in Figure~\ref{fig:all_metric_distributions}.

\section{Examples from different selection criteria}\label{sec:data_examples}
Examples from the pretraining data, drawn from distinct subsets (keep bottom, keep middle, keep top), are presented in Tables ~\ref{tab:52B_data}, \ref{tab:124M_data}, \ref{tab:wiki_data}, \ref{tab:el2n_data}, and \ref{tab:mem_data}, with rankings based on perplexity.

\begin{table*}[htb]
\small
\centering
\caption{Samples from different distribution subsets using perplexity of a 52B reference model trained on CommonCrawl.}
\label{tab:52B_data}
\begin{tabularx}{6.5in}{XXX}
\toprule
 \multicolumn{1}{c}{Bottom 10\%} &  \multicolumn{1}{c}{Middle 10\%}  &  \multicolumn{1}{c}{Top 10\%} \\
\midrule
 Submissions, you hereby grant Company a license to translate, modify (for technical purposes, for example making sure your content is viewable on an iPhone as well as a computer) and reproduce and otherwise act with respect to such User Submissions, in each case to enable us to operate the Services, as described in more detail below. This is a license only – your ownership in User Submissions is  \textcolor{lightgray}{[...]} &  House Municipal Heritage Building is a two-storey, wooden, vernacular building with a low-hipped roof, and is located at the Norris Point Lookout, 104 Main Road, Norris Point, Newfoundland and Labrador. The former family dwelling now operates as a heritage museum with a view of the Tablelands of Gros Morne National Park located on the great Northern Peninsula. The municipal heritage designation  \textcolor{lightgray}{[...]} &  and a nice book as a nice price.  Postage is via Royal Mail 1st Class in the UK. If you are buying from overseas then please contact me before completing your purchase for a quote. I will always combine P\&P so if ordering multiple books, please wait for the invoice so that discounts can be applied.  We are slowly populating our store with post war Wisden's so if there is anything you need that \textcolor{lightgray}{[...]} \\
\midrule
 provided on the Site is not intended for distribution to or use by any person or entity in any jurisdiction or country where such distribution or use would be contrary to law or regulation or which would subject us to any registration requirement within such jurisdiction or country. Accordingly, those persons who choose to access the Site from other locations do so on their own initiative and are \textcolor{lightgray}{[...]} & selection of fuel type and input of soot index, coefficient of fuel, selection of measurement units, input of date and time with keyboard and via RS232 or RS485  Procedure of industrial emissions monitoring with the use of AHKAT-410 has been agreed in FSUE "SRI Atmosphere"  AHKAT-410-16 is approved for diesel locomotive and diesel train emission monitoring at environment monitoring stations in \textcolor{lightgray}{[...]} & can be returned up to 28 days after the date of purchase.  Please note, we cannot offer refunds on beauty, pierced jewellery or on swimwear if the hygiene seal is not in place or has been broken.  We now offer FREE label-free returns with InPost Lockers (available 24/7), FREE Doddle Returns to all UK customers as well as a FREE UK Collect+ returns service via over 5,900 local stores nationwide.\textcolor{lightgray}{[...]} \\
\midrule
 license only – your ownership in User Submissions is not affected.  You agree that the licenses you grant are royalty-free, perpetual, sublicensable, irrevocable, and worldwide.  Any information or content publicly posted or privately transmitted through the Services is the sole responsibility of the person from whom such content originated, and you access all such information and content at your \textcolor{lightgray}{[...]} &  1 1/2 " steel plate, all weld construction  Hammer mill machine manufacturers, suppliers, exporters, dealers and traders in India and worldwide hammer mill machines from Gujarat and Mumbai since 1960 as per the ISO standards with required industrial features and specifications Replaceable bar type grate is available for specific applications SPECIFICATIONS :  Hammer stone crusher is a kind of equip \textcolor{lightgray}{[...]} &  several turns. Nearly a month after a foreclosure lawsuit was filed against Freestyle Music Park and its parent company, more than a dozen former department heads have sued seeking more than \$232,000 in unpaid wages and bonuses, according to court papers filed late Friday. Seventeen employees are listed as plaintiffs.  Backpay I can understand, but can you honestly expect any kind of bonuses \textcolor{lightgray}{[...]} \\
\bottomrule
\end{tabularx}
\end{table*}

\begin{table*}[htb]
\small
\centering
\caption{Samples from different distribution subsets using perplexity of a 124M reference model trained on CommonCrawl.}
\label{tab:124M_data}
\begin{tabularx}{6.5in}{XXX}
\toprule
 \multicolumn{1}{c}{Bottom 10\%} &  \multicolumn{1}{c}{Middle 10\%}  &  \multicolumn{1}{c}{Top 10\%} \\
\midrule
 risk your food going bad in a lukewarm fridge when you can lease kitchen appliances in West Hollywood through Acima!  Are you a budding DJ? A bit of a high-fidelity audio snub? Love to level up with the latest video game system? Level up your entertainment at home and on the road with sound systems for lease in West Hollywood. You can make flexible lease renewal payments on the best in-home sound \textcolor{lightgray}{[...]} &  gratitude exercise. Before you get out of bed, think of five things you are most grateful for.  If your Life Path number is 2, you have a duality fit for any earthly experience. You are deeply rooted in balance and harmony when dealing with the other numbers.  In order to stay connected to your community, start your day by connecting with your friends and family. Instead of hopping on social \textcolor{lightgray}{[...]} & keepers" definitely won't help!  Then there are those whose idea of a school librarian is based on one they remember from their childhood, who perhaps didn't let them borrow from the adult shelves or maybe told them to be quiet. You know - the cliched woman with glasses and a bun? I wear glasses myself and ended up haing to get a haircut to avoid the cliche. In summer, of course I had to put my \textcolor{lightgray}{[...]} \\
\midrule
the-art mixed-use development that features a wide variety of shops, services, and restaurants, along with over 950 luxury apartments. The sprawling urban village is pedestrian-friendly and is the perfect place if you want to indulge in a shopping spree or treat your taste buds to a hearty meal.  If you're thinking about looking for the perfect home in Brookhaven, I'm ready to help! Get in touch  \textcolor{lightgray}{[...]} &  it as a stand-alone piece but later experimented performing it as my written prediction, confabulation style, Closing Effect. It's still a work in progress but I did receive some "Standing Ovations!" ALAN ARITA  "I received a copy of GAME NIGHT and IT IS EXCELLENT! First, the quality of the book is outstanding; everything from the artwork, layout, hidden gems, and of course the precision cut \textcolor{lightgray}{[...]} &  and view the supernal beauty that lies beyond. (I wish I'd have said that first; actually I stole it from a guy who wrote it a hundred years ago!*) But if I couldn't see into the future for a few years, there wouldn't be a Christmas story today.  I've a whole lot of notes still in my jeans. One's about Rabbi Frankel of the Synagogue across West Street from old Reno High School. He was a pretty \textcolor{lightgray}{[...]} \\
\midrule
 toilet drains are overwhelmed with toilet paper or clogged by non-flushable things that find their way into the drain. If that's the case, it may be time to call a plumbing technician.  Unexpected toilet issues interrupt your daily routine, turning what you expected to be a good day right into a stressful one. You need help ASAP!  Best quality Plumbing is ready to solve your toilet troubles no \textcolor{lightgray}{[...]} &  who offer 3D printing services these days. Try searching for someone who offers them in your area.Last week, Apple announced the new A15 processor in a peculiar way: by comparing its new chip to the Android competition, rather than the A14 that powered last year's generation of iPhones. We were all left to try to infer the speed of the A15 based on Apple's claims, and wondering if the company was \textcolor{lightgray}{[...]} &  floor study, family room, kitchen, unfinished basement for future expansion \& 2 car garage. Lennar seamlessly blended \& showcased the unparalleled beauty of Colorado with the most innovative homes, energy efficient technologies \& modern conveniences, bringing the best of both worlds together. Beautiful finishes and upgrades throughout. Lennar provides the latest in energy efficiency and state of  \textcolor{lightgray}{[...]} \\
\bottomrule
\end{tabularx}
\end{table*}

\begin{table*}[htb]
\small
\centering
\caption{Samples from different distribution subsets using perplexity of a 124M reference model trained on Wikipedia.}
\label{tab:wiki_data}
\begin{tabularx}{6.5in}{XXX}
\toprule
 \multicolumn{1}{c}{Bottom 10\%} &  \multicolumn{1}{c}{Middle 10\%}  &  \multicolumn{1}{c}{Top 10\%} \\
 \midrule
 of our kids, demonstrated ability to create meaningful change, a strong commitment to learning, and an ability to work in partnership with others." Individuals accepted to this program agree to a two-year teaching commitment. If you become a core member you are required to attend an intensive summer training program to prepare for your two-year commitment. Each region has different requirements b \textcolor{lightgray}{[...]} &  HST single cylinder hydraulic cone crusher. HST single cylinder hydraulic cone crusher integrates mechanical, hydraulic, electrical, automation, intelligent control and other technologies, which can be widely used in medium, fine and ultra-fine crushing operations in metal and non-metal mines, cement, sandstone, metallurgy and other industries...  1,214 roller cone crusher products are offered \textcolor{lightgray}{[...]} & active play outdoor.  Users without a subscription are not able to see the full content on this page. Please subscribe or login.On the net betting houses include was able to offer followers a fabulous best range of luring optimistic aspects. A style of online casino money provides consistently continually really been ornamented and acquired in reaction to make sure you basic safety issues. Insi \textcolor{lightgray}{[...]} \\
\midrule
 to be that way.  Weight loss surgery in Hanover is a great option for those who are at least fifty pounds overweight and have struggled with weight loss over the years. There are a number of surgical weight loss procedures available to those seeking treatment, and Nusbaum Weight Loss Centers of New Jersey, with offices and bariatric surgeons in Morristown, Morris County, Morris County, and surrou \textcolor{lightgray}{[...]} &  sperm whales.  Learn firsthand about Sri Lanka's amazing biodiversity on this private tour to the Kanneliya Rainforest. With a dedicated guide leading you, explore the UNESCO-listed biosphere reserve, home to monkeys, snakes, chameleons, and a wide range of bird life. Learn about the flora and fauna through commentary tailored to your interests and enjoy plenty of chances to ask questions.  Explo \textcolor{lightgray}{[...]} & row for spotting this Sabal Trail posting within minutes.The skin has become delicate. I just received the goods and I didn't know how to use it. I consulted the customer service. I didn't expect the customer service person to be super good and the introduction was super careful.  I have been so successful and happy trading with you every time.. I hope we have more transactions in the future... Ha \textcolor{lightgray}{[...]} \\
\midrule
 to which coverage is thereby to be granted; and  (2) Shall insure the person named therein and any other person, as insured, using any such motor vehicle or motor vehicles with the express or implied permission of such named insured against loss from the liability imposed by law for damages arising out of the ownership, maintenance, or use of such motor vehicle or motor vehicles within the United \textcolor{lightgray}{[...]} & Also, I have attached a brief presentation of our work for better understanding.A two-year solar energy project at the University of Sheffield has shown almost all of the 2,000 systems in the scheme are still performing better than expected.  Researchers running Sheffield Solar Farm, which was launched in August 2010, say 98 per cent of more than 2,000 systems involved in the scheme are working  \textcolor{lightgray}{[...]} & It exposes a design and construction system for horizontal plates to work as slabs in regular concrete buildings. Based to an evolutionary finite-element analysis of the topological configuration to get a curved design with a 50\% reduction of traditional volume, that provide lower cost, less carbon foot-print, better performance and innovative ceiling. A library of profiles is elaborated according \textcolor{lightgray}{[...]} \\
\bottomrule
\end{tabularx}
\end{table*}

\begin{table*}[htb]
\small
\centering
\caption{Samples from different distribution subsets using EL2N from a 124M reference model trained on CommonCrawl.}
\label{tab:el2n_data}
\begin{tabularx}{6.5in}{XXX}
\toprule
 \multicolumn{1}{c}{Bottom 10\%} &  \multicolumn{1}{c}{Middle 10\%}  &  \multicolumn{1}{c}{Top 10\%} \\
\midrule
 a handle on how many elevators they are supposed to oversee.  Those officials have repeatedly deflected requests from reporters to detail the count of elevators in Chicago requiring inspection.  Frydland, during her interview, said she doesn't know how many elevators her office is responsible for inspecting because city records lump elevators into the same class of devices as escalators,  \textcolor{lightgray}{[...]} &  there's a possibility that you may come across a property that's sharing a driveway with the home next door. That means that one driveway needs to be shared between the two adjoining neighbors.  Many real estate investors rent out their properties in order to reap the benefits of passive monthly income while increasing their equity and building wealth over time. Not only are they benefiting \textcolor{lightgray}{[...]} &  We have all spent happy hours listening to and sharing music we love with those closest to us. Many of the people we serve in ubu are incredibly gifted and play a wide range of musical instruments and enjoy singing and performing for other people.  Judith is enabled by ubu to live more independently in Knaresborough, North Yorkshire, and has started taking singing lessons in order to 'grow' her \textcolor{lightgray}{[...]} \\
\midrule
ians 4:3?  Jesus addressed this very issue with his disciples on the night of his betrayal. He would be leaving them soon, but he promised the Holy Spirit would come to comfort and aide them, "I will not leave you as orphans; I will come to you."-John 14:18. Jesus refers to the Holy Spirit as himself because, "the Helper, the Holy Spirit, whom the Father will send in my name, he will teach you all \textcolor{lightgray}{[...]} &  the standard as far as cement manufacturing goes several cement manufacturers still prefer ball mills for cement production when they want to design new grinding plants or a new integrated  3D design and analysis of the crushing roller of The crushing roller is one of the main parts of a highpressure grinding roller which is a type of highly efficient ore crushing equipment In the work reported \textcolor{lightgray}{[...]} &  range (Table 1).  Active-Controlled Study: CRESTOR was compared with the HMG-CoA reductase inhibitors atorvastatin, simvastatin, and pravastatin in a multicenter, open-label, dose-ranging study of 2,240 patients with Type IIa and IIb hypercholesterolemia. After randomization, patients were treated for 6 weeks with a single daily dose of either CRESTOR, atorvastatin, simvastatin, or pravastatin \textcolor{lightgray}{[...]} \\
\midrule
 Most past attemptsto define socioeconomics as a science in its own right may have been motivated tocounter such a simplistic understanding of socioeconomics.In this chapter, we review past attempts to define socioeconomics before theapproach is chosen that we applied in this book.  This book, by a leading expert in urban agriculture, offers a genuine solution to today's global food crisis. By \textcolor{lightgray}{[...]} &  which adopted our buttons such that when we went to Boston.com (part of NY times) branding was not part of our discussions. Of course, we had matured in our thinking and offered them a co-branded offer hosted by Coola.  When Switchboard did not work for us, we went to their competition Infospace.com, which was much larger than them. They accepted a branded Coola button but offered a complex deal  \textcolor{lightgray}{[...]} & Trend.com: I had no idea this was coming. There'd been talk over the years about setting up a sort of business portal that integrated all of Trend's regular and annual publications, but there was never enough momentum to actually get it going. Trend had a regular spot on the Times' online Business section, but it was a pretty low-impact thing (even though quite a bit of traffic would come to the  \textcolor{lightgray}{[...]} \\
\bottomrule
\end{tabularx}
\end{table*}

\begin{table*}[htb]
\small
\centering
\caption{Samples from different distribution subsets using memorization of a 124M reference model trained on CommonCrawl.}
\label{tab:mem_data}
\begin{tabularx}{6.5in}{XXX}
\toprule
 \multicolumn{1}{c}{Mem. Factor = 0} &  \multicolumn{1}{c}{Mem. Factor = 0.5}  &  \multicolumn{1}{c}{Mem. Factor = 1.0} \\
\midrule
 doesn't prevent you from clearly seeing the road.  Hi, thank you so much for your words, appreciate it! Moreover, we noted your comments, we'll think what can be done, for sharing more ideas, feel free to contact us at support@hudwayapp.com any time. Happy to help you always!  I do a lot of mudding. And it's got a pitch and roll gauge, which I like when I'm in the hole, do I don't flip my truck.  \textcolor{lightgray}{[...]} &  160 countries. There are abundant hot-selling projects accessible to you.  Cheap and environmentally friendly: Factory-direct sale, fast delivery with guaranteed quality at factory price, in line with the concept of environmental development.  Feb 19 2021 should pelletisation of sulfide solidelectrolytesafterball millinghas to be done in argon atmosphere question 7 answers i am using a spex 8000b \textcolor{lightgray}{[...]} &  reference. My company's NACHI 230/600E bearing price concessions, adequate inventory, and other similar products are available for recommendation  1 . Less than 45 KGS, we will send by express. (Door to Door, Convenient) 2 . 45 - 200 KGS , we will send by air transport . (Fastest and safest, but expensive) 3 . More than 200 KGS, we will send by sea . ( Cheapest and common use )  The bearing 240/8 \textcolor{lightgray}{[...]} \\
\midrule
 disposal and processing of contaminated suspensions such as drilling mud, road sweepings and similar. The rising demand on the international market to meet current as well as future environmental regulations is the main driver for the development in this area of our work," explains Managing Director Ing. Mag. Erich Trunkenpolz. "The plants are currently developed for stationary and semi-mobile du \textcolor{lightgray}{[...]} &  \$97 monthly subscription package. If you decide to make an annual payment of \$997, you get two free months. I started with this basic package but I later decided to upgrade to Etison Suite since this one has some limitations. As a marketer, I was only allowed to use 3 custom domains, get a limit of 20,000 visitors, and make a maximum of 100 web pages. I discovered that some advanced features are  \textcolor{lightgray}{[...]} &  takes your bank to process our refund request (5 to 10 business days).  If you need to return an item, simply login to your account, view the order using the 'Complete Orders' link under the My Account menu and click the Return Item(s) button. We'll notify you via e-mail of your refund once we've received and processed the returned item.  We can ship to virtually any address in the world. Note the \textcolor{lightgray}{[...]} \\
\midrule
time:If you're looking into faster-than-light fiber internet, there's a Verizon Fios deal for you in Silver Spring, MD. Want more than a Verizon Fios internet-only plan? Open your home up to more entertainment choices with Verizon Fios packages.  Ready to improve your home with the best internet available? Get lightspeed internet with Verizon plans that suit every lifestyle. Whether you only need \textcolor{lightgray}{[...]} &   Select options that apply then copy and paste the RDF/HTML data fragment to include in your application  Note: Adjust the width and height settings defined in the RDF/HTML code fragment to best match your requirementsCause.—Upon the ascension of William and Mary to the throne of England, the Protestants of Maryland demanded the Colonial management of the Territory. The Roman Catholics, after rep \textcolor{lightgray}{[...]} &  to assess the success of our marketing and advertising campaigns).  Finally, we may also share your Personal Information to comply with applicable laws and regulations, to respond to a subpoena, search warrant or other lawful request for information we receive, or to otherwise protect our rights.  Additionally, you can opt out of some of these services by visiting the Digital Advertising Alliance \textcolor{lightgray}{[...]} \\
\bottomrule
\end{tabularx}
\end{table*}

\end{document}